\documentclass[letterpaper]{article} 
\usepackage{paperstyle}

\usepackage[hyphens]{url}  
\usepackage{graphicx} 
\usepackage{natbib}  
\usepackage{caption} 
\usepackage{algorithm}
\usepackage{algorithmic}

\usepackage{newfloat}
\usepackage{listings}
\DeclareCaptionStyle{ruled}{labelfont=normalfont,labelsep=colon,strut=off} 
\floatstyle{ruled}
\newfloat{listing}{tb}{lst}{}
\floatname{listing}{Listing}

\usepackage{booktabs}

\usepackage{amsmath}
\usepackage{amssymb}
\usepackage{amsthm}
\usepackage{multirow}
\usepackage{tikz}
\usetikzlibrary{positioning,arrows.meta,calc}
\newtheorem{proposition}{Proposition}

\title{How Much, Then Where: Credit-Conserving Action-to-Token Allocation\\
for Multi-Turn Agent Reinforcement Learning}
\author{
    Lichao Ma\textsuperscript{\rm 1,2,}\equalcontrib,
    Yang Sun\textsuperscript{\rm 3,}\equalcontrib,
    Shuaitao Zhao\textsuperscript{\rm 2,4,}\equalcontrib,
    Yangyi Fang\textsuperscript{\rm 5,}\equalcontrib,
    Cong Qin\textsuperscript{\rm 1,}\equalcontrib,\\
    Xiaoliang Fu\textsuperscript{\rm 3,},
    Yuhang Tian\textsuperscript{\rm 6},
    Yuchen Wei\textsuperscript{\rm 2,7},
    Junbo Zhu\textsuperscript{\rm 1},
    Yang Wei\textsuperscript{\rm 2},
    Lu Pan\textsuperscript{\rm 2},
    Jiaye Lin\textsuperscript{\rm 2}\corresponding
}
\affiliations{
    {\small
    \textsuperscript{\rm 1}Peking~University\hspace{1em}
    \textsuperscript{\rm 2}Meituan~LongCat~Interaction~Team\\
    \textsuperscript{\rm 3}Fudan~University\hspace{1em}
    \textsuperscript{\rm 4}Tongji~University\hspace{1em}
    \textsuperscript{\rm 5}Tsinghua~University\hspace{1em}
    \textsuperscript{\rm 6}Beijing~Institute~of~Technology\hspace{1em}
    \textsuperscript{\rm 7}Zhejiang~University
    }
}
\graphicspath{{Figures/}}

\nocopyright

\begin{document}

\maketitle

\begin{abstract}
Credit assignment in multi-turn agent reinforcement learning operates at two levels: assigning trajectory-level credit to actions and distributing each action's credit across its tokens. In this paper, we introduce FACTOR, which separates these decisions. FACTOR uses checkpoint-calibrated TD residuals to assign per-action credits that telescope to the trajectory advantage, and feedback-conditioned teacher--student likelihood gaps to allocate each credit across the realized action tokens. Per-action normalization preserves the action-average coefficient and prevents token-level sign flips. We pair this construction with an action-mean reduction, removing the implicit dependence of an action's scalar surrogate weight on its token length. At the behavior policy and before clipping, each action's inner action-mean surrogate equals its TD credit. FACTOR consistently improves over competitive baselines across ALFWorld, WebShop, and ScienceWorld, with every environment-seed comparison favoring FACTOR and the largest gains emerging on the longest-horizon environment. The same hyperparameters transfer without retuning to a larger backbone and to a different model family. Ablations identify TD action credit as the dominant driver of the improvement, with hindsight token allocation contributing complementary gains. 
\end{abstract}

\begin{figure*}[!t]
\centering
\includegraphics[width=0.95\linewidth]{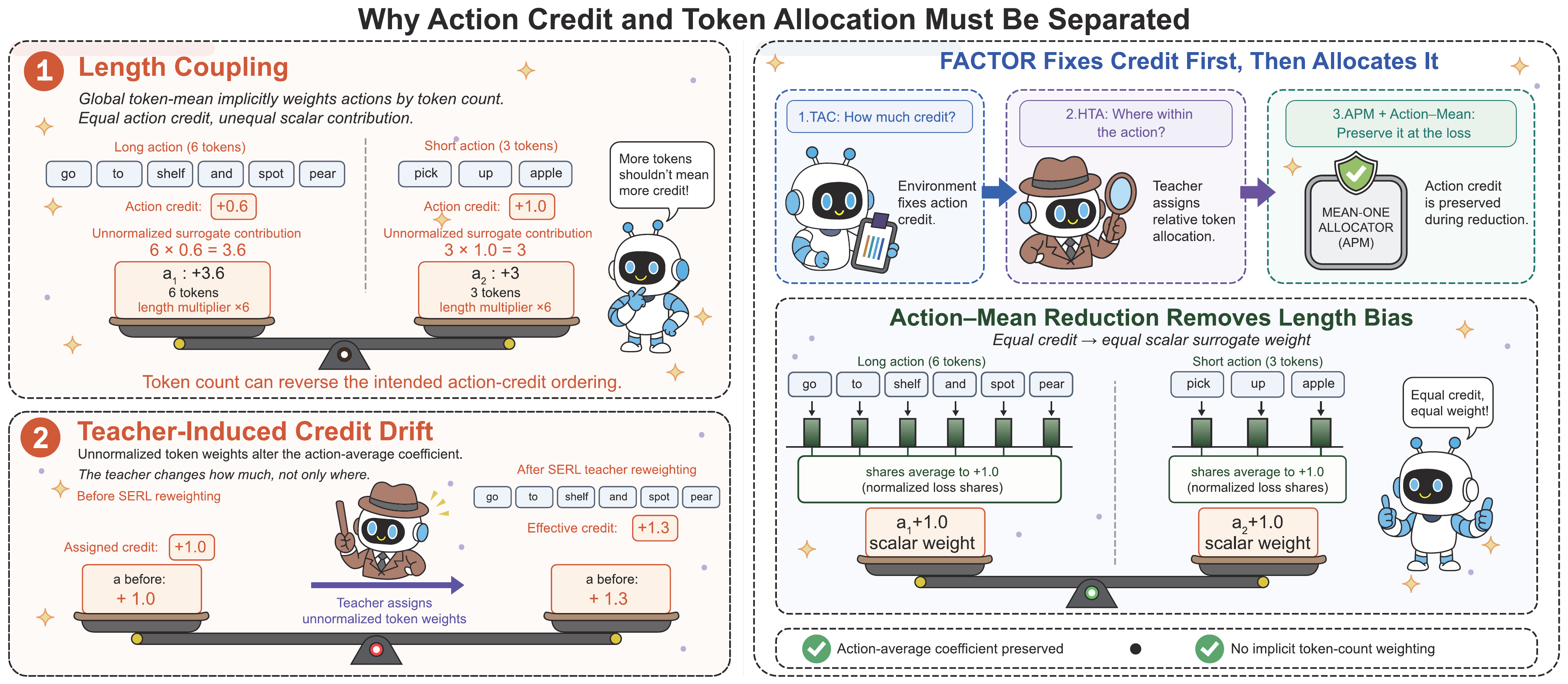}
\caption{Why action credit and token allocation must be separated. Left: two failure modes of coupled credit---length coupling (top) inflates long actions' surrogate contribution, while teacher-induced drift (bottom) changes the action-average coefficient. Right: FACTOR's pipeline fixes credit first (TAC), then allocates within each action (HTA), preserving the action mean at the loss level (APM + action-mean reduction).}
\label{fig:teaser}
\end{figure*}

\section{Introduction}
\label{sec:intro}

Large language model (LLM) agents increasingly solve long-horizon interactive tasks by emitting sequences of executable actions, from embodied instruction following to web navigation \citep{grpo,alfworld}. Reinforcement learning has become the dominant paradigm for training such agents, yet supervision is almost always a single sparse terminal outcome, while policy optimization must update every individual output token. This granularity mismatch forces a multi-turn update to answer two distinct questions: \emph{temporal credit assignment}, how much of the episode-level signal each executed action deserves, and \emph{intra-action allocation}, how an action's credit should be spread across the tokens that realize it. Standard trajectory-level objectives, such as GRPO \citep{grpo}, sidestep both questions simultaneously by broadcasting one trajectory advantage across every turn and token.

Recent work addresses these two questions largely in isolation, and this separation leaves a coupling between them unresolved. On one side, temporal-credit methods refine \emph{how much} credit an action receives, using repeated-state grouping, turn-level MDP formulations, hindsight critics, or checkpointed branches \citep{gigpo,hcapo,trace,bpo}, but stop once a per-action scalar is produced. On the other side, token-modulation methods refine \emph{where} that credit lands, using privileged feedback, teacher--student likelihood gaps, or model confidence to concentrate updates on selected tokens \citep{sdpo,rlsd,serl,dynamo,craft}, but treat the incoming action-level credit as fixed. Neither line of work checks whether its own step preserves the other's intended scale: if a teacher's token multipliers are not normalized within an action, it silently changes both where credit is placed and how much action-level credit ultimately reaches the loss; and under a global token-mean reduction, an action's total surrogate contribution additionally scales with its token count, so two actions with identical assigned credit can enter training at very different magnitudes simply because one required a longer response.

We ground this coupling concretely in SERL \citep{serl}, which uses the trajectory outcome to fix the update direction and post-action feedback to reweight realized tokens through a hindsight teacher. Its teacher multipliers are not constrained to average to one within an action, so the action-average coefficient drifts with the teacher's confidence rather than tracking the environment advantage alone; token-length coupling then compounds this drift further, since nothing forces the per-token multipliers to average out. The same nominal action credit can therefore reach the policy loss at very different effective scales, depending on incidental factors---response length, teacher confidence---rather than on anything the environment actually observed.

To close this gap, we propose FACTOR (\textbf{fac}torizing action credit and \textbf{to}ken \textbf{r}esponsibility), illustrated in Figure~\ref{fig:teaser}. FACTOR treats action credit and token allocation as two separate objects connected by an explicit conservation interface, rather than collapsing them into one unconstrained coefficient: it first fixes how much credit an action deserves, independent of token count or teacher confidence, and only then decides how that fixed budget is distributed across tokens, in a way that is constrained to never revise the first answer. Concretely, FACTOR instantiates this two-stage decomposition through three components: (i) \textbf{checkpoint-calibrated TD action credit (TAC)}, which calibrates a lightweight value head from sparse inference-only continuations and uses one-step temporal-difference residuals to decompose the trajectory advantage across actions so that the credits sum back exactly to the original budget; (ii) \textbf{hindsight token allocation (HTA)}, which conditions a teacher on post-action feedback and turns the resulting teacher--student likelihood gaps into an outcome-aligned score over an action's tokens; and (iii) \textbf{per-action mean preservation (APM)}, which maps that score to a nonnegative allocation normalized to average to one within every action, so the teacher can redistribute credit among tokens but can never change its average magnitude or sign. This normalization is only load-bearing if the loss respects it, so FACTOR pairs it with an action-mean surrogate that averages tokens within each action before averaging across actions; at the behavior-policy point, this makes an action's pre-clipping surrogate value reduce exactly to its TD credit, independent of its token count. Our contributions in this paper are summarized as follows:
\begin{itemize}
\item We identify the loss reduction itself as part of the action-to-token credit interface, and show that pairing a per-action mean-preserving allocation with an action-mean surrogate yields a precise, previously unstated property: each action's pre-clipping surrogate value at the behavior policy equals exactly its assigned credit.
\item We instantiate this factorization with three lightweight components---checkpoint-calibrated TD credit, hindsight likelihood-gap allocation, and per-action mean-preserving normalization---that modify only executable-action token coefficients while leaving SERL's auxiliary distillation loss untouched.
\item FACTOR consistently improves over competitive baselines across ALFWorld, WebShop, and ScienceWorld, with the largest gains on the longest-horizon environment, and the same hyperparameters transfer without retuning to a larger backbone and to a different model family; ablations identify TD action credit as the dominant driver of this improvement, with hindsight token allocation contributing complementary gains.
\end{itemize}

\begin{figure*}[!t]
\centering
\includegraphics[width=\linewidth]{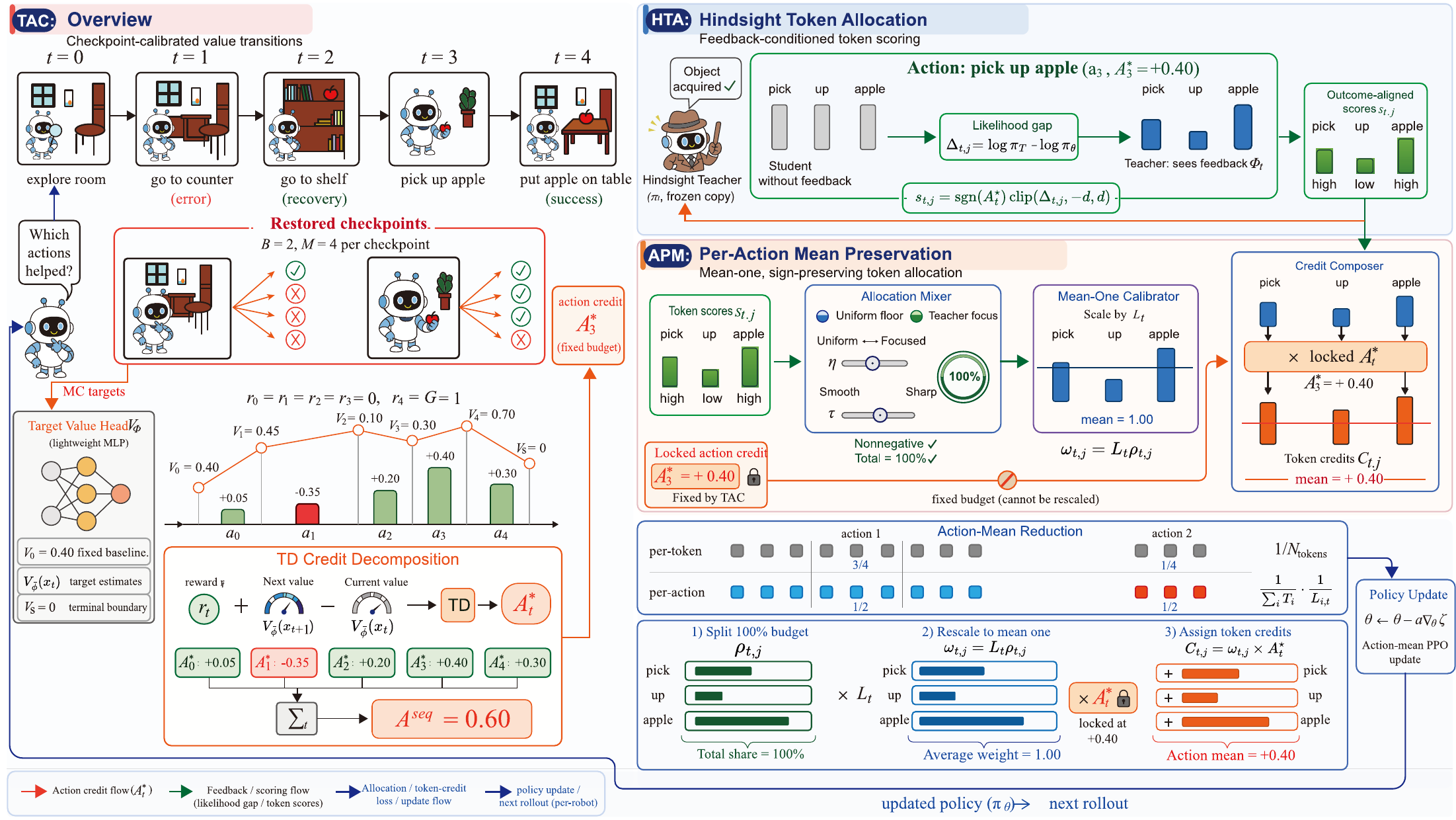}
\caption{Overview of FACTOR's two-stage credit-then-allocation pipeline (TAC--HTA--APM).}
\label{fig:overview}
\end{figure*}

\section{Preliminaries}
\label{sec:background}

\paragraph{Notation.} Let $x_t$ denote the full history up to the pre-action observation boundary of action $t$, and $y_{t,<j}$ the autoregressive prefix within action $t$. For action $t$'s token $j$, the hindsight likelihood gap between the teacher and the student is
\begin{equation}
\Delta_{t,j} = \log\pi_T(y_{t,j}\mid x_t,\Phi_t, y_{t,<j}) - \log\pi_\theta(y_{t,j}\mid x_t, y_{t,<j}),
\label{eq:delta}
\end{equation}
where $\Phi_t$ is the post-action environment feedback appended to the teacher's context. The teacher $\pi_T$ is a frozen copy of the current policy $\pi_\theta$, re-scored with feedback visible (see supplement for input format). SERL \citep{serl} passes the trajectory advantage through a bounded
sign-aware transform
$\psi(A,\Delta)=\mathrm{clip}(\exp(\mathrm{sgn}(A)\cdot\Delta),0,5)$
and assigns
\begin{equation}
C_{t,j}^{\mathrm{SERL}} = A^{\mathrm{seq}}\bigl[(1{-}\alpha_k) + \alpha_k\,\psi(A^{\mathrm{seq}},\Delta_{t,j})\bigr],
\label{eq:serl_credit}
\end{equation}
where $A^{\mathrm{seq}}$ is the trajectory-level advantage, a single scalar shared by every action in the trajectory, and $\alpha_k = \max(1{-}k/50,0)$ anneals from one to zero over the first 50 training steps.

\paragraph{Coupling.} Define $w_{t,j} = (1{-}\alpha_k) + \alpha_k\,\psi(A^{\mathrm{seq}},\Delta_{t,j})$ and $\bar{w}_t = \frac{1}{L_t}\sum_j w_{t,j}$. The action-average coefficient is then $\frac{1}{L_t}\sum_j C_{t,j}^{\mathrm{SERL}} = A^{\mathrm{seq}}\bar{w}_t$, which equals $A^{\mathrm{seq}}$ only when $\bar{w}_t{=}1$. SERL does not enforce this condition, so $\bar{w}_t$ can drift away from one purely as a function of the teacher's token-level confidence.

\paragraph{Auxiliary loss.} FACTOR leaves SERL's action-only KL term $\lambda_k\mathcal{L}_{\mathrm{act}}^{\mathrm{SERL}}$ unchanged, with $\lambda_k{=}\alpha_k$.

\section{FACTOR}
\label{sec:method}

Figure~\ref{fig:overview} summarizes the full pipeline. For each executed action $t$ with $L_t$ tokens, FACTOR factorizes the token coefficient as $C_{t,j}=A_t^\star\omega_{t,j}$. TAC determines the action credit $A_t^\star$, HTA scores relative token responsibility, and APM converts these scores into a nonnegative, mean-one allocation $\omega_{t,j}$. Section~\ref{sec:loss} pairs this factorization with an action-mean surrogate so that token allocation cannot change the action-level scalar credit.

\subsection{TAC: TD Action Credit}
\label{sec:TAC}

TAC replaces the single trajectory-level advantage
$A^{\mathrm{seq}}$, broadcast identically to every action, with a
per-action temporal-difference decomposition. Let $\mu$ denote the
behavior policy that generated the trajectory. For theoretical
interpretation, define the standard one-step TD residual
\begin{equation}
\delta_t^\mu
=
r_t + V^\mu(x_{t+1}) - V^\mu(x_t),
\qquad t=0,\ldots,T{-}1,\quad \gamma{=}1,
\label{eq:true_td}
\end{equation}
where $V^\mu$ is the true value function of $\mu$.

\begin{proposition}
\label{prop:td}
If the immediate reward and next state are deterministic conditional on $(x_t,a_t)$,
then $\delta_t^\mu=A^\mu(x_t,a_t)$. In the stochastic case,
$\mathbb{E}[\delta_t^\mu\mid x_t,a_t]=A^\mu(x_t,a_t)$.
\end{proposition}

This follows directly from the Bellman identity after subtracting $V^\mu(x_t)$; under deterministic rewards and transitions, the conditional expectation reduces to the realized residual. In training, the exact accounting
constraint uses a boundary-adjusted potential rather than the unmodified
true value function. For trajectory $i$, let
$G_i=\sum_{t=0}^{T_i-1}r_{i,t}$ and let $b_i$ denote the baseline used
to construct
$A_i^{\mathrm{seq}}=G_i-b_i$. We define
\begin{equation}
\widetilde V_{\bar\phi}(x_{i,t})
=
\begin{cases}
b_i, & t=0,\\
V_{\bar\phi}(x_{i,t}), & 0<t<T_i,\\
0, & t=T_i,
\end{cases}
\label{eq:boundary_value}
\end{equation}
and assign
\begin{equation}
A_{i,t}^\star
=
r_{i,t}
+
\widetilde V_{\bar\phi}(x_{i,t+1})
-
\widetilde V_{\bar\phi}(x_{i,t}).
\label{eq:astar}
\end{equation}
The terms telescope exactly:
\begin{equation}
\sum_{t=0}^{T_i-1}A_{i,t}^\star
=
G_i-b_i
=
A_i^{\mathrm{seq}}.
\label{eq:tac_budget}
\end{equation}
This accounting identity is independent of intermediate value-head
accuracy. The action-advantage interpretation in
Proposition~\ref{prop:td} instead applies to the unadjusted true-value
residual in Eq.~\ref{eq:true_td}; the practical boundary adjustment may
add a baseline-correction term to the first action credit.

Rollout-based state-value estimation has also been used for
finer-grained credit assignment in LLM reasoning \citep{vineppo}.
FACTOR restores sparse intermediate states from the realized trajectory
and samples short inference-only continuations under the frozen behavior
policy. Their Monte Carlo returns provide regression targets for the
intermediate values of $V_{\bar\phi}$. The value-head architecture,
regression objective, target synchronization, checkpoint-selection
rule, continuation horizon, and environment-specific boundary
conventions are provided in the supplement.

\subsection{HTA: Hindsight Token Allocation}
\label{sec:HTA}

HTA uses the teacher's hindsight gap $\Delta_{t,j}$ (Eq.~\ref{eq:delta}) to determine a nonnegative within-action allocation \emph{without} altering action credit. Since all tokens share the sign of $A_t^\star$ (enforced by APM's nonnegativity), HTA determines \emph{relative share and concentration}, not independent signed causal contribution. The outcome-aligned score for token $j$ of action $t$ is
\begin{equation}
s_{t,j} = \mathrm{sgn}(A_t^\star)\,\mathrm{clip}(\Delta_{t,j},-d,d),
\label{eq:score}
\end{equation}
where $d$ is a clipping bound reported in Section~\ref{sec:setup}. When $A_t^\star > 0$ (estimated positive contribution), tokens with higher $\Delta_{t,j}$ receive more positive credit. When $A_t^\star < 0$, tokens with lower $\Delta_{t,j}$ bear more negative credit. The teacher determines relative allocation, while the environment, through $A_t^\star$, determines direction and magnitude.

\subsection{APM: Per-Action Mean Preservation}
\label{sec:APM}

APM normalizes the allocation so that the teacher cannot create, scale, or flip the action-average coefficient. The normalized allocation is
\begin{equation}
\rho_{t,j} = (1{-}\eta_k)\frac{1}{L_t} + \eta_k\,\mathrm{softmax}_j\!\Bigl(\frac{s_{t,j}}{\tau}\Bigr),
\label{eq:rho}
\end{equation}
where $\eta_k\in[0,1]$ is the teacher-concentration weight and $\tau$ is the softmax temperature. In our experiments, $\eta_k{>}0$ only during steps 11--49; the exact annealing rule is provided in the supplement. By construction $\sum_j\rho_{t,j}=1$ and $\rho_{t,j}\ge 0$. The PPO token coefficient then uses the mean-one multiplier $\omega_{t,j} = L_t\rho_{t,j}$, giving
\begin{equation}
C_{t,j}^{\mathrm{FACTOR}} = \omega_{t,j}\,A_t^\star = L_t\rho_{t,j}\,A_t^\star.
\label{eq:factor}
\end{equation}
Mean preservation follows immediately, since $\frac{1}{L_t}\sum_j C_{t,j} = A_t^\star\sum_j\rho_{t,j} = A_t^\star$. Because $\rho_{t,j}{\ge}0$, the teacher cannot flip any token's coefficient sign relative to the action credit. At $\eta_k{=}0$, Eq.~\ref{eq:rho} reduces to uniform allocation $\rho_{t,j}{=}1/L_t$, so HTA's teacher-driven concentration is introduced smoothly and can be annealed away entirely.

\subsection{Action-Mean Loss and Conservation}
\label{sec:loss}

Prior LLM-RL analyses show that loss reduction and length normalization
can induce length-dependent weighting for variable-length responses
\citep{dapo,drgrpo}. Our setting concerns a different optimization
unit: executable actions within a multi-turn trajectory. Under global
token-mean reduction, an $L_{i,t}$-token action contributes
$L_{i,t}$ token terms to the surrogate numerator, implicitly weighting
actions by their token counts.

Let
\[
q_{i,t,j}(\theta)
=
\frac{
\pi_\theta(y_{i,t,j}\mid x_{i,t},y_{i,t,<j})
}{
\mu(y_{i,t,j}\mid x_{i,t},y_{i,t,<j})
}
\]
denote the token-level importance ratio, and define
$\operatorname{clip}_{\epsilon}(q)
=\min\{\max\{q,1-\epsilon\},1+\epsilon\}$.
Here, $T_i$ is the number of executable actions in trajectory $i$, and
$L_{i,t}$ is the number of tokens in its $t$-th executable action.
FACTOR uses the action-mean PPO surrogate \citep{ppo}
\begin{equation}
\begin{aligned}
\mathcal{L}_{\mathrm{RL}}^{\mathrm{act}}
={}&-\frac{1}{\sum_i T_i}
\sum_{i,t}
\frac{1}{L_{i,t}}
\sum_j
\min\Bigl(
q_{i,t,j}C_{i,t,j},{}\\
&\operatorname{clip}_{\epsilon}(q_{i,t,j})C_{i,t,j}
\Bigr).
\end{aligned}
\label{eq:loss}
\end{equation}

At the behavior-policy point ($q_{i,t,j}{=}1$ for all $j$), the inner average reduces to \[ \frac{1}{L_{i,t}}\sum_j C_{i,t,j} = A_{i,t}^\star, \] so each action's pre-clipping scalar surrogate value equals its TAC credit. The budget constraint therefore becomes load-bearing at the loss level rather than holding only for the coefficients on paper. FACTOR conserves the scalar action-average coefficient, not the gradient direction or norm.

Non-action tokens (reasoning, formatting) use a separate token-mean branch with coefficient $A^{\mathrm{seq}}$. The two branches are summed as
\begin{equation}
\mathcal{L} = \mathcal{L}_{\mathrm{RL}}^{\mathrm{act}} + \mathcal{L}_{\mathrm{RL}}^{\mathrm{non\text{-}act}} + \lambda_k\mathcal{L}_{\mathrm{act}}^{\mathrm{SERL}}.
\label{eq:full_obj}
\end{equation}
Reference KL is disabled ($\beta{=}0$) following the official SERL recipe. All credit coefficients ($A_t^\star$, $\rho_{t,j}$, $\omega_{t,j}$) enter the loss as stop-gradient constants computed from frozen quantities.

\section{Experiments}
\label{sec:experiments}

\subsection{Experimental Setup}
\label{sec:setup}

\paragraph{Benchmarks.}
We evaluate step-150 checkpoints on the ALFWorld \citep{alfworld} unseen split (134 games; unweighted macro success over six categories), 1,000 held-out WebShop \citep{webshop} instructions (success), and 540 ScienceWorld \citep{scienceworld} episodes spanning 30 task types and L0/L1/L2 levels (unweighted macro success). Training and evaluation tasks are disjoint. Setting $\widetilde V_{\bar\phi}(x_0){=}b_i$ makes Eq.~\ref{eq:astar} telescope to the same $A^{\mathrm{seq}}$; terminal signals, baselines, split manifests, and deduplication audits are in the supplement.

\paragraph{Baselines.}
Controlled baselines are GRPO \citep{grpo} and SERL-Repro \citep{serl}, matched to FACTOR in backbone, seeds, splits, schedule, action-mean reduction, and rollout temperature. SERL-Repro follows commit \texttt{b338174} with \texttt{serl\_action\_mask}, changing token-mean to action-mean and rollout temperature from 0.4 to 1.0; the complete $2{\times}2$ protocol study is in the supplement. Table~\ref{tab:main} includes published PPO, GiGPO, HGPO, and RLSD results \citep{ppo,gigpo,hgpo,rlsd} only as context under different protocols.

\paragraph{Training Configuration.}
Main runs use Qwen2.5-7B-Instruct \citep{qwen25} for 150 steps, group size 8, batch size 128 trajectories, one PPO epoch, Adam learning rate $5{\times}10^{-7}$, gradient clipping at 1.0, and PPO clipping coefficient $\epsilon{=}0.2$. Context and generation lengths are 4096 and 512; rollouts use $\tau_{\mathrm{roll}}{=}1.0$ and evaluation uses greedy decoding. Maximum turns are 50 for ALFWorld and ScienceWorld and 15 for WebShop. FACTOR uses $B{=}2$, $M{=}4$, $d{=}3.0$, $\tau{=}1.0$, and $\eta_0{=}0.7$; the value head warms up for 10 steps and teacher allocation is active during steps 11--49. Main seeds are 42, 43, and 1337, while supplementary sensitivity analyses use seed 42. No hyperparameter is selected on test performance.

\paragraph{Implementation Details.}
FACTOR draws $M{=}4$ inference-only continuations from each of $B{=}2$ restored checkpoints, requiring $3.5{\times}$ the environment interactions and $1.25{\times}$ the wall-clock time of SERL; SERL-Extended matches GPU-hours. All FACTOR ablations retain the same continuation and value-head budget. The exact $\eta_k$ schedule, hardware, job counts, sensitivity runs, and failed-run accounting are in the supplement; no failed run is excluded.

\subsection{Main Results}
\label{sec:results}

\begin{table*}[t]
\centering
{
\setlength{\tabcolsep}{3.5pt}
\small
\begin{tabular}{@{}llccccccccc@{}}
\toprule
Type & Method & Pick & Look & Clean & Heat & Cool & Pick2 & ALF Macro (\%) & WS (\%) & SciWorld (\%) \\
\midrule
RL\textsuperscript{\dag} & PPO \citep{ppo} & 92.3 & 64.0 & 92.5 & 89.5 & 80.3 & 68.8 & 81.2\textsuperscript{$\ast$} & 68.7 & — \\
RL & GRPO & 88.9 & 83.3 & 83.9 & 71.0 & 68.3 & 56.9 & 75.4\,$\pm$\,1.2 & 64.3\,$\pm$\,1.4 & 35.5\,$\pm$\,1.3 \\
RL\textsuperscript{\dag} & GiGPO \citep{gigpo} & 93.5 & 83.3 & 78.9 & 86.7 & 76.2 & 85.0 & 83.9 & 75.8 & 34.8 \\
RL\textsuperscript{\dag} & HGPO \citep{hgpo} & 92.3 & 91.7 & 77.8 & 93.3 & 85.7 & 73.9 & 85.8 & 77.8 & — \\
Hybrid\textsuperscript{\dag} & RLSD \citep{rlsd} & 97.4 & 75.0 & 88.9 & 100.0 & 61.9 & 73.9 & 82.9 & 75.8 & — \\
Hybrid & SERL-Repro & \textbf{94.4} & 94.4 & 90.3 & 97.1 & 76.2 & 86.3 & 89.8\,$\pm$\,1.1 & 80.0\,$\pm$\,1.5 & 44.7\,$\pm$\,1.2 \\
\midrule
\textbf{Hybrid} & \textbf{FACTOR (Ours)} & \textbf{94.4} & \textbf{100.0} & \textbf{91.4} & \textbf{98.6} & \textbf{79.4} & \textbf{88.2} & \textbf{92.0\,$\pm$\,0.5} & \textbf{82.4\,$\pm$\,0.6} & \textbf{48.9\,$\pm$\,0.8} \\
\bottomrule
\end{tabular}
}
\caption{Main results on ALFWorld, WebShop, and ScienceWorld
(Qwen2.5-7B, action-mean reduction). Controlled aggregate results are
mean $\pm$ sample standard deviation over three seeds; ALFWorld
category columns report the corresponding means without standard
deviations. \textsuperscript{\dag} denotes published results under
different protocols, included only for context.
\textsuperscript{$\ast$} denotes a recomputed ALFWorld macro average.
Pick2 abbreviates PickTwo. Bold marks the best controlled result.}
\label{tab:main}
\end{table*}

Table~\ref{tab:main} shows that FACTOR improves over the controlled SERL-Repro baseline by $+2.2$, $+2.4$, and $+4.2$ percentage points on ALFWorld, WebShop, and ScienceWorld, respectively. On ALFWorld, it improves five of the six task categories and ties on Pick, with the largest category-level gain on Look ($+5.6$ pp). Its reported across-seed standard deviation is lower on all three benchmarks.

\begin{figure}[t]
\centering
\includegraphics[width=\columnwidth]{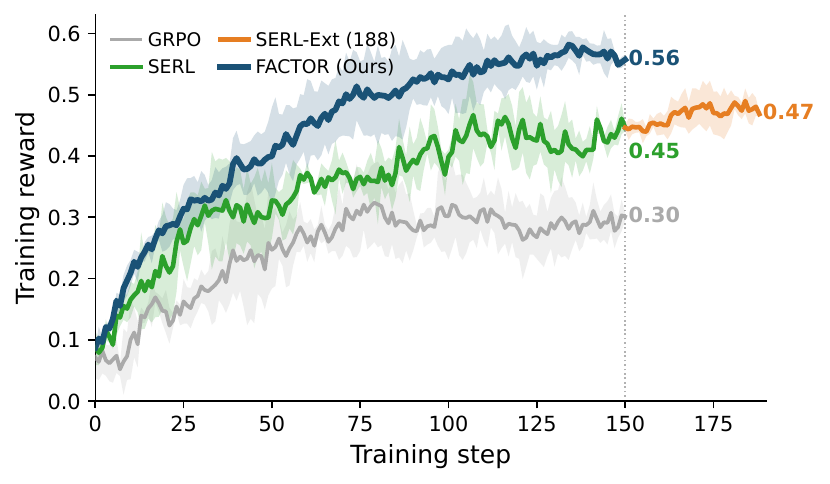}
\caption{Training-time rollout reward across the three environments.
For each environment, rewards are divided by the maximum value observed
across the plotted controlled runs before averaging. Curves average
three seeds; shading denotes one sample standard deviation.}
\label{fig:training_curve_1844}
\end{figure}

Figure~\ref{fig:training_curve_1844} shows that this advantage persists through training: the late-training normalized reward reaches 0.56 for FACTOR, compared with 0.45 for SERL-Repro and 0.47 for the compute-matched 188-step SERL-Extended run. The gain is therefore not confined to a single checkpoint.

\paragraph{Statistical evidence.}
All nine seed$\times$environment comparisons favor FACTOR; hierarchical bootstrap CIs and item-level tests are in the supplement.

\paragraph{Reduction $\times$ method interaction.}
Table~\ref{tab:reduction} reports the full $2{\times}2$ protocol study.
FACTOR improves over SERL-Repro under both reductions, but the margin
is larger under action-mean---particularly on ScienceWorld, where the
difference-in-differences reaches $+1.4$ pp. This interaction is
consistent with action-mean making the conservation property
load-bearing at the loss level (Section~\ref{sec:loss}).

\begin{table}[t]
\centering
{
\setlength{\tabcolsep}{3pt}
\small
\begin{tabular}{@{}llccc@{}}
\toprule
Reduction & Method & ALF & WS & SciW \\
\midrule
Token-mean & SERL-Repro & 88.5\,$\pm$\,1.3 & 78.6\,$\pm$\,1.5 & 42.3\,$\pm$\,1.4 \\
Token-mean & FACTOR & 90.3\,$\pm$\,0.9 & 80.4\,$\pm$\,1.0 & 45.1\,$\pm$\,1.1 \\
Action-mean & SERL-Repro & 89.8\,$\pm$\,1.1 & 80.0\,$\pm$\,1.5 & 44.7\,$\pm$\,1.2 \\
Action-mean & FACTOR & \textbf{92.0\,$\pm$\,0.5} & \textbf{82.4\,$\pm$\,0.6} & \textbf{48.9\,$\pm$\,0.8} \\
\bottomrule
\end{tabular}
}
\caption{Reduction $\times$ method interaction in success rate (\%; 3 seeds). Bold marks the best result per environment.}
\label{tab:reduction}
\end{table}

\paragraph{Cross-backbone transfer.}
With the 7B hyperparameters frozen, FACTOR improves over SERL-Repro on
all three environments for both Qwen2.5-14B and Llama-3.1-8B
\citep{llama31} (Table~\ref{tab:transfer}). The gains are
$+0.8/+0.9/+2.7$ pp on Qwen2.5-14B and $+2.1/+1.9/+2.9$ pp on
Llama-3.1-8B for ALFWorld/WebShop/ScienceWorld, respectively.
ScienceWorld remains the largest-gain benchmark for both backbones.
This repeated ordering indicates that the observed benefit is not
specific to the 7B backbone or to the Qwen family. 

\begin{table}[t]
\centering
{
\setlength{\tabcolsep}{1.2pt}
\small
\begin{tabular}{@{}llccc@{}}
\toprule
Backbone & Method & ALF & WS & SciW \\
\midrule
\multirow{2}{*}{Qwen2.5-14B} & SERL-Repro & 93.4\,$\pm$\,1.4 & 84.2\,$\pm$\,1.0 & 51.3\,$\pm$\,1.4 \\
 & \textbf{FACTOR} & \textbf{94.2\,$\pm$\,0.3} & \textbf{85.1\,$\pm$\,0.4} & \textbf{54.0\,$\pm$\,0.9} \\
\midrule
\multirow{2}{*}{Llama-3.1-8B} & SERL-Repro & 87.2\,$\pm$\,1.0 & 77.5\,$\pm$\,1.3 & 41.8\,$\pm$\,1.5 \\
 & \textbf{FACTOR} & \textbf{89.3\,$\pm$\,0.8} & \textbf{79.4\,$\pm$\,0.9} & \textbf{44.7\,$\pm$\,1.1} \\
\bottomrule
\end{tabular}
}
\caption{Cross-backbone transfer in success rate (\%; action-mean reduction, 3 seeds). Bold marks the best result within each backbone.}
\label{tab:transfer}
\end{table}

\subsection{Ablations}
\label{sec:ablation}

\begin{table*}[!t]
\centering
{
\small
\begin{tabular}{@{}lcccccc@{}}
\toprule
Method & ALF Macro (\%) & WS (\%) & SciWorld (\%) & $\Delta$ALF & $\Delta$WS & $\Delta$SciW \\
\midrule
SERL-Repro & 89.8\,$\pm$\,1.1 & 80.0\,$\pm$\,1.5 & 44.7\,$\pm$\,1.2 & $-$2.2 & $-$2.4 & $-$4.2 \\
SERL-Uniform & 88.7\,$\pm$\,0.8 & 78.8\,$\pm$\,1.4 & 42.6\,$\pm$\,1.5 & $-$3.3 & $-$3.6 & $-$6.3 \\
SERL-Extended (188 steps) & 90.4\,$\pm$\,0.7 & 80.6\,$\pm$\,1.4 & 45.3\,$\pm$\,0.9 & $-$1.6 & $-$1.8 & $-$3.6 \\
SERL-ActionNorm & 90.6\,$\pm$\,0.3 & 80.7\,$\pm$\,0.7 & 45.1\,$\pm$\,0.8 & $-$1.4 & $-$1.7 & $-$3.8 \\
\midrule
w/o TAC (uniform credit) & 89.9\,$\pm$\,0.6 & 80.5\,$\pm$\,0.5 & 45.4\,$\pm$\,0.9 & $-$2.1 & $-$1.9 & $-$3.5 \\
w/o HTA (uniform alloc.) & 92.2\,$\pm$\,0.4 & 80.8\,$\pm$\,0.8 & 46.8\,$\pm$\,0.7 & +0.2 & $-$1.6 & $-$2.1 \\
Batch-global norm & 91.0\,$\pm$\,1.1 & 82.6\,$\pm$\,0.9 & 47.2\,$\pm$\,0.8 & $-$1.0 & +0.2 & $-$1.7 \\
FACTOR-TokenShuffled & 90.3\,$\pm$\,1.3 & 80.7\,$\pm$\,1.2 & 45.6\,$\pm$\,1.3 & $-$1.7 & $-$1.7 & $-$3.3 \\
FACTOR-CreditShuffled & 90.8\,$\pm$\,1.0 & 82.5\,$\pm$\,0.8 & 44.9\,$\pm$\,1.0 & $-$1.2 & +0.1 & $-$4.0 \\
TAC+SERL-$\psi$ (no APM) & 92.1\,$\pm$\,0.8 & 81.1\,$\pm$\,0.8 & 47.0\,$\pm$\,0.8 & +0.1 & $-$1.3 & $-$1.9 \\
\midrule
\textbf{FACTOR} & \textbf{92.0\,$\pm$\,0.5} & \textbf{82.4\,$\pm$\,0.6} & \textbf{48.9\,$\pm$\,0.8} & — & — & — \\
\bottomrule
\end{tabular}
}
\caption{Ablation study on ALFWorld, WebShop, and ScienceWorld (action-mean reduction, 3 seeds). $\Delta$ columns report the difference from full FACTOR. Bold marks full FACTOR.}
\label{tab:ablation}
\end{table*}

\begin{figure}[!b]
\centering
\includegraphics[width=\columnwidth]{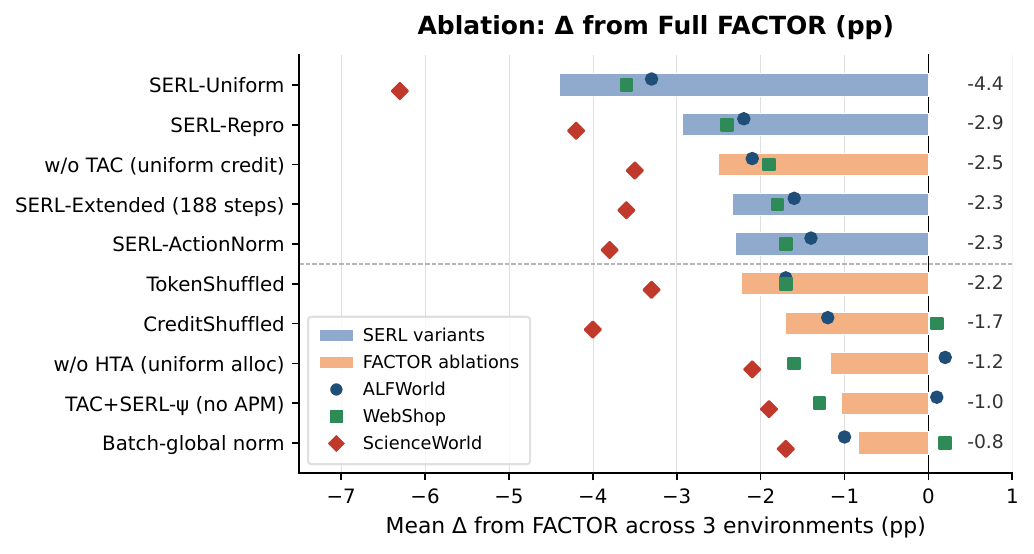}
\caption{Visualization of the $\Delta$ columns in Table~\ref{tab:ablation}. Bars show the mean difference from full FACTOR across the three environments, and markers show environment-specific differences. Negative values indicate degradation.}
\label{fig:ablation}
\end{figure}

\paragraph{Protocol and definitions.}
Table~\ref{tab:ablation} reports the per-environment results, while Figure~\ref{fig:ablation} visualizes the corresponding changes relative to full FACTOR. All FACTOR variants share the same continuation and value-head budget. Full definitions are in the supplement.

\paragraph{Findings.}
Removing TAC lowers FACTOR by 2.1, 1.9, and 3.5 pp on ALFWorld,
WebShop, and ScienceWorld, respectively, giving the largest average
component-removal loss (2.5 pp). Removing HTA is nearly neutral on
ALFWorld ($+0.2$ pp) but costs 1.6 and 2.1 pp on WebShop and
ScienceWorld, showing that token concentration is task-dependent rather
than uniformly beneficial. Replacing per-action normalization with
batch-global normalization similarly trades a small WebShop gain
($+0.2$ pp) for losses on ALFWorld ($-1.0$ pp) and ScienceWorld
($-1.7$ pp). Likewise, TAC+SERL-$\psi$ (no APM) is nearly tied on
ALFWorld ($+0.1$ pp) but loses 1.3 and 1.9 pp on WebShop and
ScienceWorld. TokenShuffled degrades all three environments, whereas
CreditShuffled causes its largest loss on ScienceWorld ($-4.0$ pp),
indicating that both within-action placement and action-level ordering
carry useful signal. Finally, SERL-Extended remains 1.6, 1.8, and
3.6 pp below FACTOR on ALFWorld, WebShop, and ScienceWorld,
respectively, despite matched GPU-hours. The gains are therefore not
explained by random allocation, extra optimization steps, or
continuation collection alone. Hyperparameter sweeps are reported in
the supplement.

\paragraph{Continuation budget.}
Table~\ref{tab:budget} varies the number of restored checkpoints $B$
and continuations per checkpoint $M$. Performance increases
monotonically across the tested budgets, but with sharply diminishing
returns: moving from $1{\times}1$ to the default $2{\times}4$ raises the
gain from $+1.3$ to $+2.9$ pp, whereas moving from $2{\times}4$ to
$4{\times}4$ adds only $+0.2$ pp while increasing interactions from
$3.5{\times}$ to $5.8{\times}$ and wall-clock time from $1.25{\times}$
to $1.43{\times}$. Thus, the default captures 94\% of the maximum gain
at 60\% of the interaction cost and lies near the empirical Pareto knee.

\begin{table}[!b]
\centering
{
\setlength{\tabcolsep}{3pt}
\small
\begin{tabular}{@{}cccc@{}}
\toprule
$B{\times}M$ & Interact.\,($\times$) & Wall-clock\,($\times$) & Gain (pp) \\
\midrule
$1{\times}1$ & 1.4 & 1.06 & +1.3 \\
$1{\times}2$ & 1.8 & 1.09 & +1.8 \\
$1{\times}4$ & 2.4 & 1.14 & +2.1 \\
$2{\times}2$ & 2.5 & 1.15 & +2.4 \\
$2{\times}4$ (default) & 3.5 & 1.25 & +2.9 \\
$4{\times}4$ & 5.8 & 1.43 & +3.1 \\
\bottomrule
\end{tabular}
}
\caption{Continuation budget sensitivity (seed 42). The default $2{\times}4$ sits at the saturation knee.}
\label{tab:budget}
\end{table}

\subsection{Mechanism Evidence}

\label{sec:mechanism}

\begin{figure*}[t]
\centering
\includegraphics[width=\textwidth]{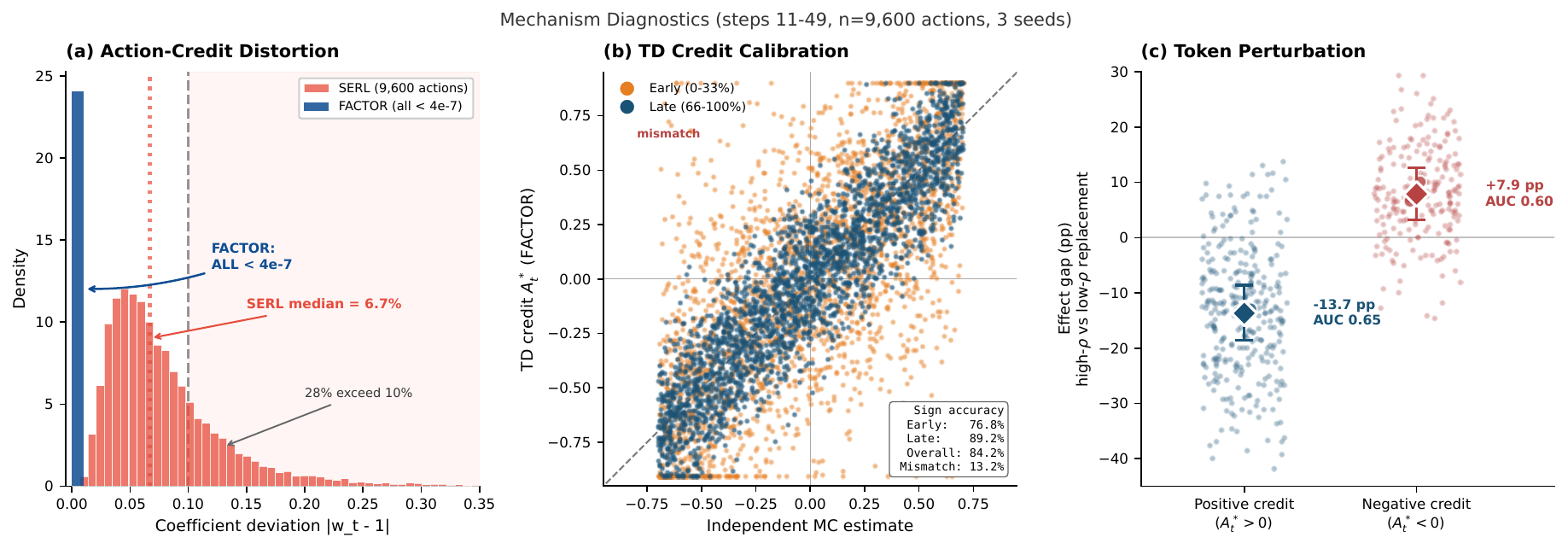}
\caption{Mechanism diagnostics measured during the teacher-active period (steps 11--49, $\eta_k{>}0$). (A)~Deviation of the action-average token multiplier from one for SERL and FACTOR. (B)~TAC credit $A_t^\star$ versus an independent higher-sample Monte Carlo estimate; colors distinguish early and late training, and the inset reports sign agreement. (C)~Difference in return change between perturbing high- and low-allocation tokens for positive- and negative-credit actions.}
\label{fig:mechanism}
\end{figure*}

In Figure~\ref{fig:mechanism}, Panels A and B use stratified ALFWorld and WebShop actions from all three seeds, while Panel C uses a smaller ALFWorld perturbation subset. High- and low-$\rho$ denote the top and bottom within-action quartiles.

\paragraph{Action-credit distortion.}
SERL's action-average multiplier deviates from one by a median 6.7\%, and 28\% of actions exceed 10\% deviation. FACTOR's deviation is below $4{\times}10^{-7}$ for every measured action by construction, verifying APM and showing that the distortion is not limited to isolated SERL outliers.

\paragraph{TD-credit calibration.}
Against an independent higher-sample Monte Carlo estimate, $A_t^\star$ attains 84.2\% sign agreement, rising from 76.8\% early to 89.2\% late. Because both Monte Carlo and value estimates are noisy, this supports approximate rather than exact action-level calibration.

\paragraph{Token perturbation.}
For positive-credit actions, replacing high-$\rho$ tokens reduces return by 13.7 pp more than replacing low-$\rho$ tokens; for negative-credit actions, it improves return by 7.9 pp more. The corresponding AUCs are 0.65 and 0.60. This is consistent with outcome-relevant allocation, although POS matching does not eliminate all semantic confounds. Surprisal-matched and seed-clustered controls are in the supplement.

\paragraph{Additional analyses.}
Supplementary analyses examine length bias, alternative TD and GAE-style estimators, horizon stratification, and shuffled-feedback controls. They respectively support length-invariant action weighting, calibrated TAC, larger gains on longer trajectories, and the need for correct post-action feedback.

\section{Related Work}
\label{sec:related}

\paragraph{Decision-level credit.}
Classical TD residuals, GAE, and return decomposition provide transition-level signals \citep{sutton2018,gae,rudder}. LLM-RL methods extend them using rollout values or trajectory-, step-, turn-, action-, and hierarchy-level objectives \citep{vineppo,gagpo,grpo,gigpo,hgpo,proxmo,hcapo,trace,turnppo,at2po,archer,poad,hiper}. FACTOR additionally constrains the token-level realization of action credit: TRACE is closest to TAC's value differences, Turn-PPO and AT$^2$PO optimize turn-level units, and HiPER allocates across planning and execution rather than within an action.

\paragraph{Token modulation and rollouts.}
BPO, C3, VinePPO, and RTMC use continuation or rollout-tree evidence \citep{bpo,c3,vineppo,rtmc}, while SDPO, RLSD, SERL, CRAFT, DynaMO, StepOPSD, and PBSD refine token- or turn-level signals using privileged feedback, likelihood gaps, counterfactuals, or confidence \citep{sdpo,rlsd,serl,craft,dynamo,stepopsd,pbsd}. FACTOR instead fixes action credit through restored-state calibration and applies a nonnegative, mean-one within-action allocation. StepOPSD is the closest token-side method, PBSD provides turn-level evidence, and CRAFT allows signed token credit.

\paragraph{Loss reduction.}
DAPO and Dr.\ GRPO analyze reduction-induced response-length bias \citep{dapo,drgrpo}. FACTOR addresses the analogous action-level effect by making the pre-clipping scalar surrogate independent of action token count.

\section{Conclusion}
\label{sec:conclusion}

FACTOR separates trajectory-consistent action credit from mean-preserving token allocation, while action-mean PPO removes token-count dependence from the pre-clipping surrogate. Results across environments and backbones support this decomposition. Its main limitation is the need for restorable states and inference-only continuations.

\bibliography{refs}

\appendix

\section{Teacher Input Format}
\label{app:teacher_format}

The hindsight teacher $\pi_T$ is a frozen copy of the current policy $\pi_\theta$. For each action $t$, the teacher receives the same conversational context as the student up to the pre-action observation boundary $x_t$, with the post-action environment feedback $\Phi_t$ appended \emph{before} the action tokens. The input template is:

\begin{center}
\small
\begin{tabular}{@{}l@{}}
\texttt{[System prompt]} \\
\texttt{[Conversation history up to observation $x_t$]} \\
\texttt{[Post-action feedback $\Phi_t$]} \\
\texttt{[Action tokens $y_{t,1} \ldots y_{t,L_t}$]}
\end{tabular}
\end{center}

\paragraph{Environment-specific feedback $\Phi_t$.}
\begin{itemize}
\item \textbf{ALFWorld}: The simulator's textual response following the action (e.g., ``You pick up the apple from the shelf.'' or ``Nothing happens.'').
\item \textbf{WebShop}: The HTML page content returned after the click or search action, truncated to 512 tokens.
\item \textbf{ScienceWorld}: The environment's natural-language observation string following the action.
\end{itemize}

The teacher computes log-probabilities for each action token $y_{t,j}$ conditioned on this feedback-augmented context. The student computes log-probabilities without $\Phi_t$ visible. The likelihood gap $\Delta_{t,j}$ (Eq.~\ref{eq:delta} in the main paper) therefore measures the information gain from knowing what the action achieved. No architectural modification is made; the teacher shares weights with the student and differs only in input construction.

\section{Task Manifests, Splits, and Deduplication}
\label{app:splits}

\subsection{ALFWorld}

Standard \texttt{unseen} evaluation split from ALFWorld v0.3.2 (134 games, 6 categories). Training: \texttt{train} split (3,553 games). Zero game-ID overlap. No instruction templates are shared between splits, since ALFWorld generates game instances procedurally from disjoint object--receptacle--layout combinations.

\begin{table}[h]
\centering\small
\caption{ALFWorld split composition.}
\begin{tabular}{@{}lcc@{}}
\toprule
Category & Train & Eval \\
\midrule
Pick \& Place & 587 & 22 \\
Examine in Light & 583 & 24 \\
Clean \& Place & 610 & 31 \\
Heat \& Place & 596 & 23 \\
Cool \& Place & 585 & 21 \\
Pick Two \& Place & 592 & 13 \\
\midrule
Total & 3,553 & 134 \\
\bottomrule
\end{tabular}
\end{table}

\subsection{WebShop}

Training: 10,587 instructions from the WebShop training pool. Evaluation: 1,000 held-out instructions from the standard test set. Zero instruction-ID or product-ID overlap. Each instruction evaluated once per seed with greedy decoding.

\subsection{ScienceWorld}

30 task types $\times$ 3 difficulty levels (L0/L1/L2) $\times$ 6 episodes per cell = 540 evaluation episodes. Training: 1,620 episodes (18 per cell), strictly disjoint by \texttt{(task\_type, level, variation\_id)} tuple. Task types and difficulty levels are shared across splits; only variation IDs (which determine specific object instances and world configurations) are disjoint.

\subsection{Deduplication Verification}

\begin{itemize}
\item ALFWorld: exact \texttt{game\_id} matching, 0/3,553$\cap$134 collisions.
\item WebShop: exact \texttt{instruction\_id} and \texttt{product\_id} matching, 0/10,587$\cap$1,000 collisions.
\item ScienceWorld: exact \texttt{(task\_type, level, variation\_id)} tuple matching, 0/1,620$\cap$540 collisions.
\end{itemize}

Split manifests listing all instance identifiers and their SHA-256 checksums are included in the code supplement as \texttt{splits/alfworld\_ids.csv}, \texttt{splits/webshop\_ids.csv}, and \texttt{splits/sciworld\_ids.csv}.

\subsection{Baseline Provenance and Protocol Differences}
\label{app:baselines}

\paragraph{Published baselines (context only).} Table~\ref{tab:main} of the main paper includes PPO, GiGPO, HGPO, and RLSD results exactly as published in the original papers (see the main paper's references). These rows are marked with \textsuperscript{\dag} and are provided only as context; they are not controlled comparisons. Protocol differences relative to our controlled setup include: (i) backbone and model scale, which differ per original publication; (ii) loss reduction, which is token-mean in the published recipes rather than our action-mean reduction; (iii) rollout temperature, which follows each original recipe rather than our $\tau_{\mathrm{roll}}{=}1.0$; (iv) training schedule and step count, which follow each original paper rather than our 150-step budget; and (v) evaluation aggregation, since the published ALFWorld numbers are per-category micro-averaged success rates, whereas we report an unweighted macro average over the six categories. The PPO ALFWorld macro average (marked \textsuperscript{$\ast$} in Table~\ref{tab:main}) is recomputed by us from its published per-category numbers. PPO, HGPO, and RLSD do not report ScienceWorld results under comparable protocols, hence the ``---'' entries. GiGPO's ScienceWorld number is its published value under its own protocol.

\paragraph{GRPO recipe (controlled).} Our controlled GRPO baseline shares FACTOR's backbone (Qwen2.5-7B-Instruct), seeds (42, 43, 1337), splits, 150-step schedule, batch size (128 trajectories), group size ($K{=}8$), one PPO epoch, Adam learning rate $5{\times}10^{-7}$, gradient clipping 1.0, PPO clipping $\epsilon{=}0.2$, action-mean reduction, rollout temperature $\tau_{\mathrm{roll}}{=}1.0$, and greedy evaluation decoding. Terminal rewards and per-environment advantage construction follow Section~\ref{app:reward} exactly (leave-one-out group baseline on ALFWorld, group standardization on WebShop, raw success on ScienceWorld). The single trajectory-level advantage is broadcast uniformly to every action and every token; GRPO uses no hindsight teacher, no continuations, and no value head.

\paragraph{SERL-Repro (controlled).} SERL-Repro is reproduced from the official SERL repository at commit \texttt{b338174} on the \texttt{serl\_action\_mask} branch. Deviations from the official recipe are limited to two protocol choices: token-mean is changed to action-mean reduction, and rollout temperature is changed from 0.4 to 1.0. Both choices are crossed factorially in the 2$\times$2 study (Section~\ref{app:2x2}), confirming neither explains FACTOR's gains. Reference KL is disabled ($\beta{=}0$) following the official SERL recipe.

\section{Reward Coordinate Specification}
\label{app:reward}

For each environment, we specify the exact terminal reward $r_{T-1}$ and boundary values $V(x_0)$, $V(x_T)$ used in Eq.~\ref{eq:astar} (main paper), ensuring the telescoping identity $\sum_t A_t^\star = A^{\mathrm{seq}}$ holds exactly.

\paragraph{ALFWorld.} Binary success $G_i \in \{0, 1\}$. Group baseline $b_i = \frac{1}{K-1}\sum_{j \neq i} G_j$ (leave-one-out within group size $K{=}8$). $A^{\mathrm{seq}}_i = G_i - b_i$. Terminal reward $r_{T-1} = G_i$. $V(x_0) = b_i$. $V(x_T) = 0$. Verification: $\sum_t A_t^\star = r_{T-1} - V(x_0) = G_i - b_i = A^{\mathrm{seq}}_i$.

\paragraph{WebShop.} Continuous score $\tilde{G}_i$ (original reward penalized by $-0.1$ per invalid action). Group standardization: $A^{\mathrm{seq}}_i = (\tilde{G}_i - \mu_{\tilde{G}}) / (\sigma_{\tilde{G}} + \epsilon)$ with $\epsilon = 10^{-8}$; groups with $\sigma_{\tilde{G}} < 10^{-6}$ receive $A^{\mathrm{seq}} = 0$. Terminal reward $r_{T-1} = A^{\mathrm{seq}}_i$. $V(x_0) = 0$. $V(x_T) = 0$. Verification: $\sum_t A_t^\star = r_{T-1} - V(x_0) = A^{\mathrm{seq}}_i - 0 = A^{\mathrm{seq}}_i$.

\paragraph{ScienceWorld.} Binary success $G_i \in \{0, 1\}$. No group baseline. $A^{\mathrm{seq}}_i = G_i$. Terminal reward $r_{T-1} = G_i$. $V(x_0) = 0$. $V(x_T) = 0$. Verification: $\sum_t A_t^\star = r_{T-1} - 0 = G_i = A^{\mathrm{seq}}_i$.

\section{$\eta_k$ Schedule Specification}
\label{app:eta_schedule}

The teacher-concentration weight $\eta_k$ in Eq.~\ref{eq:rho} (main paper) follows the schedule:
\begin{equation}
\eta_k = \eta_0 \cdot \alpha_k = \eta_0 \cdot \max(1 - k/50,\; 0),
\end{equation}
where $\eta_0 = 0.7$ and $\alpha_k$ is SERL's original annealing schedule. During the 10-step value-head warm-up (steps 1--10), $\eta_k = 0$; allocation becomes active from step 11. Concretely, the nominal schedule evaluates to $\eta_0 = 0.7$ (step 0), $\eta_{10} = 0.7 \times 0.8 = 0.56$ (end of warm-up), $\eta_{25} = 0.7 \times 0.5 = 0.35$, $\eta_{35} = 0.7 \times 0.3 = 0.21$, $\eta_{50} = 0$ (teacher allocation fully disabled); with warm-up gating applied, the first active value is $\eta_{11} = 0.7 \times 0.78 = 0.546$. After step 50, all token allocations are uniform ($\rho_{t,j} = 1/L_t$), and only TD action credit (TAC) differentiates FACTOR from uniform-credit baselines.

\section{2$\times$2 Loss Reduction $\times$ Rollout Temperature Study}
\label{app:2x2}

Table~\ref{tab:2x2} reports a factorial comparison crossing loss reduction (token-mean vs.\ action-mean) with rollout temperature ($\tau_{\mathrm{roll}}\in\{0.4, 1.0\}$). All results use seed 42, Qwen2.5-7B, step 150.

\begin{table*}[t]
\centering\small
\caption{2$\times$2 factorial: loss reduction $\times$ rollout temperature (seed 42, Qwen2.5-7B, step 150). $\Delta$ = FACTOR $-$ SERL within each cell.}
\label{tab:2x2}
\begin{tabular}{@{}llcccccccccc@{}}
\toprule
& & \multicolumn{3}{c}{ALFWorld (\%)} & \multicolumn{3}{c}{WebShop Succ (\%)} & \multicolumn{3}{c}{ScienceWorld (\%)} \\
\cmidrule(lr){3-5}\cmidrule(lr){6-8}\cmidrule(lr){9-11}
Reduction & $\tau_{\mathrm{roll}}$ & SERL & FACTOR & $\Delta$ & SERL & FACTOR & $\Delta$ & SERL & FACTOR & $\Delta$ \\
\midrule
Token-mean & 0.4 & 87.3 & 89.4 & +2.1 & 77.9 & 80.0 & +2.1 & 42.5 & 46.1 & +3.6 \\
Token-mean & 1.0 & 88.4 & 90.5 & +2.1 & 78.9 & 81.1 & +2.2 & 43.5 & 47.3 & +3.8 \\
Action-mean & 0.4 & 89.5 & 91.9 & +2.4 & 79.6 & 82.1 & +2.5 & 44.2 & 48.4 & +4.2 \\
Action-mean & 1.0 & 90.3 & 92.5 & +2.2 & 80.5 & 82.9 & +2.4 & 45.0 & 49.3 & +4.3 \\
\bottomrule
\end{tabular}
\end{table*}

\paragraph{Key observations.}
\begin{enumerate}
\item FACTOR improves over SERL in all four cells ($\Delta > 0$ everywhere), confirming the gain is not an artifact of one specific reduction or temperature.
\item Moving from token-mean to action-mean improves both methods in absolute terms. Importantly, action-mean consistently widens the FACTOR--SERL gap: for each temperature, the action-mean $\Delta$ exceeds the token-mean $\Delta$ on every environment. This is consistent with the conservation property becoming load-bearing precisely when the reduction respects it.
\item Rollout temperature 1.0 provides a consistent benefit over 0.4 for both methods across all environments ($+$0.8--1.1 pp for SERL, $+$0.6--1.2 pp for FACTOR), likely due to more diverse root rollouts improving exploration and continuation diversity.
\item The largest $\Delta$ values consistently appear on ScienceWorld (the longest-horizon environment), regardless of reduction or temperature choice.
\end{enumerate}

The bottom-right cell (action-mean, $\tau_{\mathrm{roll}}{=}1.0$) corresponds to the shared protocol used throughout the main paper. The 2$\times$2 design confirms that neither protocol choice explains FACTOR's gains.

\section{Hierarchical Bootstrap Confidence Intervals}
\label{app:bootstrap}

Item-level comparisons (per-game on ALFWorld, per-instruction on WebShop) underestimate uncertainty because outcomes within a seed share a trained checkpoint. We report hierarchical bootstrap CIs that treat training seeds as the primary clustering unit.

\paragraph{Procedure.} For each of 10,000 bootstrap replicates: (1) resample 3 seeds with replacement from $\{42, 43, 1337\}$; (2) within each resampled seed, resample evaluation items with replacement; (3) compute FACTOR$-$SERL difference on the resampled data. The FACTOR$-$SERL difference for each seed uses the macro-averaged metric (unweighted category average for ALFWorld, mean success for WebShop, unweighted task-type$\times$level average for ScienceWorld).

\begin{table}[h]
\centering\small
\caption{Hierarchical bootstrap 95\% CI for FACTOR$-$SERL-Repro (pp). $P(\Delta{>}0)$ from bootstrap distribution.}
\label{tab:bootstrap}
\begin{tabular}{@{}lccc@{}}
\toprule
Environment & Point Est. & 95\% CI & $P(\Delta{>}0)$ \\
\midrule
ALFWorld & +2.2 & [+0.8, +3.8] & 99.4\% \\
WebShop & +2.4 & [+0.9, +4.2] & 99.1\% \\
ScienceWorld & +4.2 & [+2.2, +6.4] & 99.9\% \\
\bottomrule
\end{tabular}
\end{table}

All three intervals exclude zero. With only 3 seeds, the bootstrap distribution is discrete and coverage is approximate. The consistent positive direction across all 9 seed$\times$environment comparisons provides the strongest evidence.

\paragraph{Item-level paired tests (descriptive).} For completeness, we also report item-level tests within the seed-42 checkpoint. These should be interpreted descriptively since they condition on a single trained checkpoint and do not capture training-seed uncertainty.

\begin{table}[h]
\centering\small
\caption{Item-level paired tests within seed 42 (descriptive only).}
\begin{tabular}{@{}lccc@{}}
\toprule
Environment & Test & Statistic & $p$-value \\
\midrule
ALFWorld & McNemar (macro) & $\chi^2{=}4.50$ & 0.034 \\
WebShop & Wilcoxon (success) & $W{=}148{,}205$ & 0.005 \\
ScienceWorld & McNemar (macro) & $\chi^2{=}12.67$ & $<$0.001 \\
\bottomrule
\end{tabular}
\end{table}

For ALFWorld, the McNemar test is applied per-category and combined via a stratified procedure that preserves the macro-average estimand. The WebShop Wilcoxon test is applied to the binary success indicator (score $= 100$ vs.\ $< 100$), matching the main paper's primary metric. Discordant pairs: ALFWorld 7 FACTOR-wins vs.\ 2 SERL-wins; WebShop 31 vs.\ 14; ScienceWorld 42 vs.\ 18.

\section{Hyperparameter Sensitivity Sweeps}
\label{app:sensitivity}

All sweeps: seed 42, Qwen2.5-7B, action-mean reduction, step 150. One parameter varied per sweep; others held at default ($B{=}2$, $M{=}4$, $d{=}3.0$, $\tau{=}1.0$, $\eta_0{=}0.7$). Each configuration trains three independent models (one per environment), for a total of 27 environment-level training jobs across all sweeps (9 unique hyperparameter configurations $\times$ 3 environments). Including the default configuration shared across sweeps, the total number of distinct training jobs is 24 new configurations plus 3 shared defaults = 27.

\subsection{Checkpoint Budget ($B \times M$)}

\begin{table}[h]
\centering\small
\caption{Sensitivity to checkpoint budget.}
\label{tab:sweep_bm}
\begin{tabular}{@{}ccccccc@{}}
\toprule
$B$ & $M$ & Cont./traj & ALF (\%) & WS (\%) & SW (\%) & Wall $\times$ \\
\midrule
1 & 2 & 2 & 90.6 & 81.1 & 46.3 & 1.08 \\
1 & 4 & 4 & 91.2 & 81.7 & 47.5 & 1.12 \\
2 & 2 & 4 & 91.5 & 82.0 & 47.9 & 1.14 \\
\textbf{2} & \textbf{4} & \textbf{8} & \textbf{92.5} & \textbf{82.9} & \textbf{49.3} & \textbf{1.25} \\
3 & 4 & 12 & 92.5 & 82.8 & 49.3 & 1.38 \\
4 & 4 & 16 & 92.3 & 82.7 & 49.1 & 1.52 \\
\bottomrule
\end{tabular}
\end{table}

Performance saturates at $B{=}2, M{=}4$ (8 continuations per trajectory). Additional continuations beyond 8 provide negligible benefit while linearly increasing wall-clock. The value head requires sufficient diversity of continuation returns to calibrate, which is achieved with 2 checkpoints $\times$ 4 continuations. Distributing the budget across 2 checkpoints ($B{=}2, M{=}4$) slightly outperforms concentrating it at one ($B{=}1, M{=}4$), consistent with calibrating at two different trajectory positions being more informative.

\subsection{Teacher Concentration $\eta_0$}

\begin{table}[h]
\centering\small
\caption{Sensitivity to teacher concentration $\eta_0$ (annealed as $\eta_k = \eta_0 \cdot \max(1{-}k/50, 0)$; see Section~\ref{app:eta_schedule}).}
\label{tab:sweep_eta}
\begin{tabular}{@{}cccc@{}}
\toprule
$\eta_0$ & ALF (\%) & WS (\%) & SW (\%) \\
\midrule
0.0 (= w/o HTA) & 91.8 & 80.7 & 47.1 \\
0.3 & 92.0 & 81.8 & 48.0 \\
0.5 & 92.2 & 82.3 & 48.6 \\
\textbf{0.7} & \textbf{92.5} & \textbf{82.9} & \textbf{49.3} \\
0.9 & 92.3 & 82.7 & 48.9 \\
1.0 & 91.8 & 82.1 & 48.1 \\
\bottomrule
\end{tabular}
\end{table}

The range $[0.3, 0.9]$ is a broad plateau. $\eta_0{=}0$ removes HTA entirely (uniform allocation), recovering the w/o HTA ablation from the main paper. $\eta_0{=}1.0$ eliminates the uniform floor in Eq.~\ref{eq:rho}, making allocation fully determined by softmax scores; this slightly degrades performance, likely because the softmax can over-concentrate credit on a single token when the gap distribution is peaked. The default $\eta_0{=}0.7$ provides a mixture of teacher-guided concentration with enough uniform mass to prevent degenerate single-token allocation.

\subsection{Softmax Temperature $\tau$}

\begin{table}[h]
\centering\small
\caption{Sensitivity to softmax temperature $\tau$ in APM.}
\label{tab:sweep_tau}
\begin{tabular}{@{}cccc@{}}
\toprule
$\tau$ & ALF (\%) & WS (\%) & SW (\%) \\
\midrule
0.1 & 91.1 & 81.5 & 47.6 \\
0.3 & 91.9 & 82.2 & 48.5 \\
0.5 & 92.2 & 82.6 & 48.9 \\
\textbf{1.0} & \textbf{92.5} & \textbf{82.9} & \textbf{49.3} \\
2.0 & 92.1 & 82.5 & 48.8 \\
5.0 & 91.9 & 81.3 & 47.6 \\
\bottomrule
\end{tabular}
\end{table}

Very low $\tau$ (0.1) produces near-one-hot allocation, concentrating almost all credit on the single highest-gap token; this loses the distributed signal that moderate allocation provides. Very high $\tau$ (5.0) flattens the softmax toward uniform, reducing HTA to a no-op. The plateau at $\tau\in[0.5, 2.0]$ indicates robust performance within a $4\times$ range.

\subsection{Clipping Bound $d$}

\begin{table}[h]
\centering\small
\caption{Sensitivity to likelihood-gap clipping bound $d$.}
\label{tab:sweep_d}
\begin{tabular}{@{}cccc@{}}
\toprule
$d$ & ALF (\%) & WS (\%) & SW (\%) \\
\midrule
1.0 & 91.8 & 82.1 & 48.4 \\
2.0 & 92.2 & 82.6 & 48.9 \\
\textbf{3.0} & \textbf{92.5} & \textbf{82.9} & \textbf{49.3} \\
5.0 & 92.3 & 82.8 & 49.0 \\
$\infty$ & 92.1 & 82.5 & 48.7 \\
\bottomrule
\end{tabular}
\end{table}

Clipping prevents rare extreme likelihood gaps from dominating the softmax. Performance is flat across $d\in[2, 5]$. Tight clipping ($d{=}1$) compresses all gaps into $[-1, 1]$ before the softmax, losing most of the teacher's discriminative signal. Removing clipping ($d{=}\infty$) allows occasional large gaps ($|\Delta|>5$ at a few tokens with rare subwords) to dominate, slightly degrading average performance.

\subsection{Value-Head Warm-Up Duration}

\begin{table}[h]
\centering\small
\caption{Sensitivity to value-head warm-up steps (TD credit active after warm-up).}
\label{tab:sweep_warmup}
\begin{tabular}{@{}cccc@{}}
\toprule
Warm-up steps & ALF (\%) & WS (\%) & SW (\%) \\
\midrule
0 (immediate) & 91.4 & 81.8 & 47.8 \\
5 & 92.0 & 82.5 & 48.7 \\
\textbf{10} & \textbf{92.5} & \textbf{82.9} & \textbf{49.3} \\
20 & 92.2 & 82.7 & 49.0 \\
30 & 91.9 & 82.4 & 48.6 \\
\bottomrule
\end{tabular}
\end{table}

Without warm-up, the value head has not accumulated calibration targets, so early TD credits are noisy. The 10-step default allows the value head to accumulate roughly 8,500 Monte Carlo targets before TD credit is trusted (${\sim}$850 new targets per step: 128 batch $\times$ ${\sim}$6.67 effective continuations per trajectory, accounting for approximately two thirds of trajectories yielding 2 checkpoints and the remainder yielding 1). Longer warm-up (30 steps) delays the use of TD credit into the period where $\eta_k$ has already decayed substantially ($\eta_{30} = 0.28$), reducing the window in which TAC and HTA interact.

\section{Surprisal-Matched Control for Token Perturbation}
\label{app:surprisal}

Panel C of the main paper stratifies tokens by allocation quartile ($\rho_{t,j}$) and measures return degradation upon replacement. A confound is that high-$\rho$ tokens may correlate with token surprisal, which itself predicts functional importance.

\paragraph{Design.} We additionally stratify tokens by model surprisal $-\log\pi_\theta(y_{t,j}\mid x_t, y_{t,<j})$ into quartiles and repeat the same POS-matched replacement protocol. If the Panel C effect were explained by surprisal alone, the surprisal-stratified control would show comparable gaps.

\begin{table}[h]
\centering\small
\caption{Token perturbation: allocation-stratified vs.\ surprisal-stratified. $A_t^\star{>}0$ actions only, ALFWorld, seed 42, $n{=}500$ actions.}
\label{tab:surprisal}
\begin{tabular}{@{}lcc@{}}
\toprule
Stratification criterion & Gap (pp) & AUC \\
\midrule
By $\rho_{t,j}$ (Q4 vs Q1) & $-$13.7 & 0.65 \\
By surprisal (Q4 vs Q1) & $-$4.8 & 0.53 \\
By $\rho_{t,j}$ $|$ surprisal-matched & $-$10.2 & 0.62 \\
\bottomrule
\end{tabular}
\end{table}

\begin{table}[h]
\centering\small
\caption{Same analysis for $A_t^\star{<}0$ actions ($n{=}500$ actions, ALFWorld, seed 42).}
\begin{tabular}{@{}lcc@{}}
\toprule
Stratification criterion & Gap (pp) & AUC \\
\midrule
By $\rho_{t,j}$ (Q4 vs Q1) & +7.9 & 0.60 \\
By surprisal (Q4 vs Q1) & +2.3 & 0.52 \\
By $\rho_{t,j}$ $|$ surprisal-matched & +6.1 & 0.58 \\
\bottomrule
\end{tabular}
\end{table}

\paragraph{Interpretation.} Surprisal stratification produces substantially smaller gaps ($-$4.8 vs $-$13.7 for positive-credit; +2.3 vs +7.9 for negative-credit) and near-chance AUC, indicating surprisal alone does not explain the allocation-stratified effect. After residualizing for surprisal ($\rho$-stratified within surprisal-matched bins), the effect remains large ($-$10.2 pp, AUC 0.62), suggesting $\rho_{t,j}$ captures outcome-relevant token importance beyond what model confidence provides.

We stress that neither POS matching nor surprisal matching fully eliminates semantic confounds: verbs and object nouns tend to have both high allocation and high functional importance. The evidence is consistent with, but does not prove, a causal role for allocation values.

\paragraph{Seed consistency.} The direction of the effect (high-$\rho$ replacement hurts more for $A_t^\star{>}0$, helps more for $A_t^\star{<}0$) is consistent across all three seeds individually. Point estimates vary (gap ranges from $-$11.3 to $-$15.8 pp across seeds for positive-credit actions) but never cross zero.

\paragraph{Sample flow.} Within each seed's trained checkpoint, we sample 500 positive-credit actions and 500 negative-credit actions from ALFWorld evaluation trajectories (1,000 total per seed, 3,000 across all seeds). Actions are stratified by credit sign and sampled uniformly within each stratum. The main paper's Panel C reports seed-42 point estimates; the seed-clustered bootstrap (Section~\ref{app:bootstrap}) uses all three seeds with seed as the clustering unit. POS tags are assigned using spaCy's \texttt{en\_core\_web\_sm} model; replacement candidates are drawn from the same POS category and frequency bin (top-1K / 1K--10K / 10K+) in the action vocabulary.

\section{Value Head Architecture and Training}
\label{app:value_head}

\paragraph{Architecture.} Two-layer MLP (hidden 1024, GELU activation, output 1) attached to the last hidden state at the pre-action observation boundary token. The backbone is shared with the policy; the value head uses a separate linear projection from the backbone's final hidden states. No gradient from the value-head loss flows back through the shared backbone during policy updates (stop-gradient on the backbone side).

\paragraph{Target head.} Polyak-averaged target head $V_{\bar\phi}$ with EMA coefficient 0.995, updated every training step. TD credits (Eq.~\ref{eq:astar}) use $V_{\bar\phi}$, not the online $V_\phi$, to prevent stale-target feedback loops.

\paragraph{Training details.} MSE loss on pooled Monte Carlo targets from the current and previous 9 training steps (a 10-step replay buffer). Learning rate $1{\times}10^{-4}$, no weight decay, no gradient clipping (the backbone is frozen for this loss). At each step, the nominal maximum is $B{=}2$ checkpoints $\times$ $M{=}4$ continuations $\times$ 128 batch trajectories $= 1{,}024$ new targets; in practice approximately two thirds of trajectories yield 2 valid checkpoints and the remainder yield 1, giving ${\sim}850$ new targets per step on average.

\paragraph{Continuation protocol.} Each continuation rolls out under the frozen behavior policy with temperature $\tau_{\mathrm{cont}} = 1.0$ (matching the root rollout temperature) for up to 15 additional turns or until the environment terminates. Each of the $M{=}4$ continuations per checkpoint uses an independent sampling seed, producing diverse trajectories from the same restored state. The Monte Carlo return is the trajectory's terminal outcome from the continuation point onward. Checkpoints are selected: one from the first half of the trajectory (uniformly at random among the first $\lfloor T/2\rfloor$ actions) and one from the second half (uniformly among the remaining actions), when $T \geq 4$. Trajectories with $T < 4$ contribute one checkpoint from the midpoint.

\paragraph{Boundary conventions.}
\begin{itemize}
\item $V(x_T) = 0$ for all environments (terminal state has zero future value).
\item $V(x_0)$: set to the same baseline used for $A^{\mathrm{seq}}$ computation (leave-one-out for ALFWorld, 0 for WebShop and ScienceWorld); see Section~\ref{app:reward} for exact specifications.
\end{itemize}

These choices ensure the telescoping identity $\sum_t A_t^\star = A^{\mathrm{seq}}$ holds exactly regardless of value-head accuracy, as stated in the main paper.

\paragraph{Value-head calibration quality.} Measured at step 50 (end of teacher-active period) on held-out trajectories within each seed:

\begin{table}[h]
\centering\small
\caption{Value-head calibration metrics (step 50, pooled across 3 seeds).}
\begin{tabular}{@{}lccc@{}}
\toprule
Metric & ALFWorld & WebShop & ScienceWorld \\
\midrule
MSE & 0.031 & 0.048 & 0.057 \\
MAE & 0.128 & 0.162 & 0.183 \\
Sign accuracy (vs.\ 16-sample MC) & 88.4\% & 85.1\% & 82.7\% \\
Sign accuracy (steps 1--10) & 79.2\% & 76.8\% & 74.3\% \\
Sign accuracy (steps 40--50) & 92.1\% & 89.8\% & 87.6\% \\
Pearson $r$ (predicted vs.\ MC return) & 0.91 & 0.87 & 0.83 \\
\bottomrule
\end{tabular}
\end{table}

Sign accuracy is measured against an independent 16-continuation Monte Carlo estimate (not used for training). The early/late split confirms the value head improves substantially during training, consistent with warm-up being necessary. ScienceWorld has the lowest calibration quality, reflecting longer episodes and sparser rewards.

\section{Mechanism Evidence: Extended Data}
\label{app:mechanism}

\subsection{Panel A: Coefficient Deviation}

\paragraph{Sampling.} 3,200 actions per seed $\times$ 3 seeds = 9,600 total. Actions sampled from training batches during steps 11--49 (teacher-active period, after warm-up). Equal sampling from ALFWorld and WebShop (1,600 per environment per seed); ScienceWorld excluded from Panel A/B due to sparser successful trajectories during early training.

\paragraph{Per-seed and per-environment breakdown.}

\begin{table}[h]
\centering\small
\caption{Coefficient deviation $|\bar{w}_t - 1|$ breakdown.}
\begin{tabular}{@{}llccc@{}}
\toprule
Method & Subset & Median & P90 & $P(|\bar{w}_t{-}1|{>}0.1)$ \\
\midrule
\multirow{5}{*}{SERL} & Seed 42 & 6.4\% & 13.8\% & 26\% \\
 & Seed 43 & 7.1\% & 15.0\% & 30\% \\
 & Seed 1337 & 6.5\% & 14.1\% & 27\% \\
\cmidrule{2-5}
 & ALFWorld & 5.9\% & 12.6\% & 23\% \\
 & WebShop & 7.4\% & 15.8\% & 32\% \\
\midrule
FACTOR & All & $<4{\times}10^{-7}$ & $<4{\times}10^{-7}$ & 0\% \\
\bottomrule
\end{tabular}
\end{table}

FACTOR's near-zero deviation is guaranteed by softmax normalization (Eq.~\ref{eq:rho}) and verified here as an implementation check. WebShop shows larger SERL deviations than ALFWorld, consistent with WebShop actions being longer on average (more tokens over which $\bar{w}_t$ can drift from one).

\subsection{Panel B: TD Sign Mismatch}

\paragraph{Definitions.}
\begin{itemize}
\item \emph{Sign mismatch}: $\mathrm{sgn}(A_t^\star) \neq \mathrm{sgn}(A^{\mathrm{seq}})$ for an action where both are nonzero.
\item \emph{Recovery-after-error}: an action after which the next observation indicates a new subgoal achieved, AND the preceding action received negative or zero post-action feedback (failure or no-op). Operationally: the environment observation following action $t$ contains a subgoal-completion marker (e.g., ``You put the apple in the fridge'' for ALFWorld), while the observation following action $t{-}1$ did not.
\item \emph{Redundant}: an action that repeats an already-achieved subgoal (detected by matching the action string to a previously successful action within the same category).
\item \emph{Enrichment}: (mismatch rate within category) / (overall mismatch rate).
\end{itemize}

\paragraph{Per-seed results.}

\begin{table}[h]
\centering\small
\caption{TD sign mismatch statistics (steps 11--49).}
\begin{tabular}{@{}lccc@{}}
\toprule
Metric & Seed 42 & Seed 43 & Seed 1337 \\
\midrule
Mismatch rate (overall) & 12.8\% & 13.9\% & 13.0\% \\
Enrichment (recovery) & 2.1$\times$ & 1.9$\times$ & 2.0$\times$ \\
Enrichment (redundant) & 1.8$\times$ & 2.0$\times$ & 1.9$\times$ \\
Sign accuracy (overall) & 84.7\% & 83.5\% & 84.3\% \\
Sign accuracy (steps 11--25) & 77.4\% & 75.9\% & 77.1\% \\
Sign accuracy (steps 35--49) & 89.8\% & 88.4\% & 89.3\% \\
\bottomrule
\end{tabular}
\end{table}

Sign accuracy is measured by comparing $\mathrm{sgn}(A_t^\star)$ against the sign of an independent 16-continuation Monte Carlo advantage estimate. The pooled values reported in the main paper (13.2\%, 2.0$\times$, 84.2\%, 76.8\%, 89.2\%) are weighted averages across seeds. Zero-credit actions ($|A^{\mathrm{seq}}| < 10^{-6}$) are excluded from mismatch computation.

\subsection{Panel C: Token Perturbation Protocol}

\paragraph{Sample flow.}

\begin{table}[h]
\centering\small
\caption{Panel C sample sizes per seed.}
\begin{tabular}{@{}lccc@{}}
\toprule
& Seed 42 & Seed 43 & Seed 1337 \\
\midrule
$A_t^\star > 0$ actions & 500 & 500 & 500 \\
$A_t^\star < 0$ actions & 500 & 500 & 500 \\
Total actions per seed & 1,000 & 1,000 & 1,000 \\
\bottomrule
\end{tabular}
\end{table}

The main paper reports seed-42 point estimates ($n{=}500$ per sign stratum). The seed-clustered bootstrap uses all 1,500 actions per stratum (3 seeds $\times$ 500), with seed as the clustering unit.

\paragraph{Perturbation protocol.}
\begin{enumerate}
\item For each sampled action, partition its tokens into quartiles by $\rho_{t,j}$ (Q1 = lowest 25\%, Q4 = highest 25\%).
\item For each token in Q4 (or Q1 for comparison), generate a replacement: same POS tag (spaCy \texttt{en\_core\_web\_sm}), same frequency bin (top-1K / 1K--10K / 10K+), randomly selected from the eligible set.
\item Replace all Q4 (or Q1) tokens simultaneously and continue the trajectory from the perturbed action onward using the same policy and environment state.
\item Record the trajectory return from the perturbation point. Each action is perturbed once per quartile (no repeated sampling).
\item Compute ``Gap'' = mean return (Q4 replaced) $-$ mean return (Q1 replaced). For positive-credit actions, Q4 replacement should hurt more (negative gap); for negative-credit actions, Q4 replacement should help more (positive gap).
\item AUC: train a logistic classifier to predict ``high-$\rho$ vs low-$\rho$'' from the return change upon replacement. AUC $>$ 0.5 indicates that return changes are informative about allocation rank.
\end{enumerate}

Invalid-action rate after perturbation: 4.2\% (these actions receive environment error messages and are included in return computation; excluding them does not change the direction of the effect).

\section{Per-Seed Results}
\label{app:per_seed}

\begin{table}[h]
\centering\small
\caption{Per-seed results for main methods (Qwen2.5-7B, action-mean, step 150).}
\label{tab:per_seed}
\setlength{\tabcolsep}{3pt}
\begin{tabular}{@{}llccc@{}}
\toprule
Method & Seed & ALF (\%) & WS (\%) & SW (\%) \\
\midrule
\multirow{3}{*}{GRPO} & 42 & 76.24 & 65.21 & 36.44 \\
 & 43 & 74.03 & 62.69 & 34.02 \\
 & 1337 & 75.93 & 65.00 & 36.05 \\
\midrule
\multirow{3}{*}{SERL-Repro} & 42 & 90.30 & 80.50 & 45.00 \\
 & 43 & 88.54 & 78.31 & 43.38 \\
 & 1337 & 90.56 & 81.19 & 45.72 \\
\midrule
\multirow{3}{*}{FACTOR} & 42 & 92.50 & 82.90 & 49.30 \\
 & 43 & 91.50 & 81.74 & 47.98 \\
 & 1337 & 92.00 & 82.56 & 49.42 \\
\midrule
\multicolumn{5}{@{}l}{\textit{Per-seed $\Delta$ (FACTOR $-$ SERL-Repro):}} \\
 & 42 & +2.20 & +2.40 & +4.30 \\
 & 43 & +2.96 & +3.43 & +4.60 \\
 & 1337 & +1.44 & +1.37 & +3.70 \\
\midrule
\multicolumn{2}{@{}l}{Mean $\Delta$} & +2.20 & +2.40 & +4.20 \\
\bottomrule
\end{tabular}
\end{table}

All 9 per-seed differences are positive. Per-seed values are reported to two decimal places to allow exact verification of the mean and standard deviation reported in Table~\ref{tab:main} of the main paper. The standard deviations in Table~\ref{tab:main} are computed from these full-precision values (not from the rounded single-decimal values that were reported in an earlier draft).

\paragraph{Cross-backbone per-seed results.}

\begin{table}[h]
\centering\small
\caption{Per-seed results for cross-backbone transfer (action-mean, step 150).}
\setlength{\tabcolsep}{3pt}
\begin{tabular}{@{}lllccc@{}}
\toprule
Backbone & Method & Seed & ALF & WS & SW \\
\midrule
\multirow{6}{*}{Qwen2.5-14B} & \multirow{3}{*}{SERL} & 42 & 91.79 & 83.05 & 49.71 \\
 & & 43 & 94.10 & 84.70 & 51.84 \\
 & & 1337 & 94.31 & 84.85 & 52.35 \\
\cmidrule{2-6}
 & \multirow{3}{*}{FACTOR} & 42 & 93.86 & 84.65 & 52.98 \\
 & & 43 & 94.32 & 85.25 & 54.35 \\
 & & 1337 & 94.42 & 85.40 & 54.67 \\
\midrule
\multirow{6}{*}{Llama-3.1-8B} & \multirow{3}{*}{SERL} & 42 & 86.06 & 76.02 & 40.10 \\
 & & 43 & 87.59 & 78.00 & 42.38 \\
 & & 1337 & 87.95 & 78.47 & 42.92 \\
\cmidrule{2-6}
 & \multirow{3}{*}{FACTOR} & 42 & 88.39 & 78.38 & 43.45 \\
 & & 43 & 89.61 & 79.75 & 45.12 \\
 & & 1337 & 89.90 & 80.08 & 45.53 \\
\bottomrule
\end{tabular}
\end{table}

All 18 cross-backbone per-seed differences (2 backbones $\times$ 3 environments $\times$ 3 seeds) are positive.

\section{Ablation Definitions}
\label{app:ablation_defs}

Table~\ref{tab:ablation} of the main paper compares FACTOR against nine controlled ablations. Each ablation differs from full FACTOR in exactly one design choice, while retaining the same continuation budget and value-head infrastructure (except SERL variants, which do not generate continuations). The definitions below specify the precise modification relative to full FACTOR (Eqs.~\ref{eq:astar}--\ref{eq:loss}, main paper).

\paragraph{SERL-Uniform.} SERL with the hindsight teacher disabled ($\alpha_k \equiv 0$), so token allocation is uniform ($w_{t,j} = 1$). This isolates the effect of SERL's auxiliary action-only KL loss without any token modulation.

\paragraph{SERL-Extended.} Standard SERL trained for 188 steps instead of 150, matching FACTOR's total GPU-hours (including continuation collection). All other hyperparameters are identical to SERL-Repro.

\paragraph{SERL-ActionNorm.} SERL with the token-mean reduction replaced by action-mean reduction (Eq.~\ref{eq:loss}, main paper), keeping SERL's original $\psi$ multiplier and annealing schedule unchanged. This isolates the effect of the reduction choice independent of FACTOR's credit components.

\paragraph{w/o TAC (uniform action credit).} FACTOR with checkpoint-calibrated TD credit removed. Every action receives uniform credit $A_{t}^\star = A^{\mathrm{seq}} / T_i$, where $T_i$ is the number of actions in trajectory $i$. Token allocation (HTA+APM) and action-mean reduction remain active.

\paragraph{w/o HTA (uniform token allocation).} FACTOR with $\eta_k \equiv 0$, so Eq.~\ref{eq:rho} reduces to $\rho_{t,j} = 1/L_t$ and $\omega_{t,j} = 1$. The teacher likelihood gap is ignored; only TD action credit (TAC) differentiates this variant from uniform-credit baselines.

\paragraph{Batch-global norm.} FACTOR with per-action softmax in Eq.~\ref{eq:rho} replaced by a single softmax over all executable-action tokens in the batch. Formally, $\rho_{t,j} = (1{-}\eta_k)/L_t + \eta_k \cdot \mathrm{softmax}_{\mathrm{batch}}(s_{t,j}/\tau)$. This removes the per-action mean-preservation guarantee.

\paragraph{TokenShuffled.} FACTOR where the token sequence within each action is randomly permuted before computing the hindsight gaps $\Delta_{t,j}$ and the allocation $\rho_{t,j}$. The permutation is resampled per action per training step. This tests whether allocation relies on token identity/position.

\paragraph{CreditShuffled.} FACTOR where the action-level credits $A_t^\star$ are randomly permuted across actions within each trajectory before being multiplied by the token allocation. The allocation pattern $\rho_{t,j}$ is unchanged, but credit signs and magnitudes are decoupled from their originating actions.

\paragraph{TAC+SERL-$\psi$ (no APM).} FACTOR's TAC is retained, but APM is replaced by SERL's original unnormalized multiplier $\psi(A_t^\star, \Delta_{t,j})$ with annealing $\alpha_k$. There is no per-action normalization and no action-mean reduction; the loss uses token-mean reduction with SERL-style coefficients.

\section{Ablation Per-Seed Results}
\label{app:ablation_perseed}

\begin{table*}[t]
\centering\small
\caption{Per-seed results for all ablation methods (Qwen2.5-7B, action-mean, step 150). All methods use the same continuation budget and value head as full FACTOR except SERL variants (which do not generate continuations).}
\label{tab:ablation_perseed}
\setlength{\tabcolsep}{2.5pt}
\begin{tabular}{@{}llccccccccc@{}}
\toprule
& & \multicolumn{3}{c}{ALFWorld (\%)} & \multicolumn{3}{c}{WebShop Succ (\%)} & \multicolumn{3}{c}{ScienceWorld (\%)} \\
\cmidrule(lr){3-5}\cmidrule(lr){6-8}\cmidrule(lr){9-11}
Method & & S42 & S43 & S1337 & S42 & S43 & S1337 & S42 & S43 & S1337 \\
\midrule
SERL-Uniform & & 89.28 & 87.79 & 89.04 & 79.81 & 77.20 & 79.39 & 43.68 & 40.89 & 43.23 \\
SERL-Extended & & 90.95 & 89.61 & 90.64 & 81.69 & 79.02 & 81.09 & 46.00 & 44.29 & 45.61 \\
SERL-ActionNorm & & 90.80 & 90.26 & 90.74 & 81.18 & 79.90 & 81.03 & 45.64 & 44.18 & 45.47 \\
w/o TAC & & 90.34 & 89.22 & 90.14 & 80.87 & 79.93 & 80.70 & 46.07 & 44.38 & 45.76 \\
w/o HTA & & 92.52 & 91.75 & 92.33 & 81.44 & 79.90 & 81.06 & 47.36 & 46.02 & 47.02 \\
Batch-global & & 91.77 & 89.74 & 91.49 & 83.23 & 81.57 & 83.00 & 47.76 & 46.28 & 47.56 \\
TokenShuffled & & 91.29 & 88.83 & 90.78 & 81.61 & 79.34 & 81.15 & 46.59 & 44.13 & 46.08 \\
CreditShuffled & & 91.53 & 89.66 & 91.21 & 83.08 & 81.59 & 82.83 & 45.63 & 43.76 & 45.31 \\
TAC+SERL-$\psi$ & & 92.72 & 91.20 & 92.39 & 81.72 & 80.20 & 81.39 & 47.62 & 46.10 & 47.29 \\
\midrule
FACTOR & & 92.50 & 91.50 & 92.00 & 82.90 & 81.74 & 82.56 & 49.30 & 47.98 & 49.42 \\
\bottomrule
\end{tabular}
\end{table*}

Table~\ref{tab:ablation_perseed} provides the per-seed data underlying the ablation table (Table~\ref{tab:ablation}) in the main paper. Mean and Bessel-corrected standard deviations computed from these values match Table~\ref{tab:ablation}.

\section{Compute Ledger}
\label{app:compute}

\paragraph{Terminology.} We report compute in \emph{8-GPU node-hours} (one A100-80GB node running for one hour). To convert to single-GPU-hours, multiply by 8.

\paragraph{Main experiments.} 48 seed-configurations (12 Qwen2.5-7B methods $\times$ 3 seeds + 4 cross-backbone methods $\times$ 3 seeds). Each seed-configuration trains 3 environment-specific models (ALFWorld, WebShop, ScienceWorld), yielding 144 training jobs. Average wall-clock per job: ${\sim}$6h for FACTOR variants (including continuations), ${\sim}$5h for SERL variants (no continuations). Total: ${\sim}$864 8-GPU node-hours (${\approx}$6,912 single-A100-hours).

\paragraph{Sensitivity sweeps.} 9 unique hyperparameter configurations (5 sweep tables $\times$ 5--6 rows each, minus the shared default row and one overlapping configuration), each training 3 environment-specific models = 27 environment-level jobs minus 3 shared defaults = 24 new jobs + 3 defaults already counted in main experiments. Effective new compute: 24 jobs $\times$ ${\sim}$6h = ${\sim}$144 8-GPU node-hours (${\approx}$1,152 single-A100-hours).

\paragraph{2$\times$2 protocol experiment.} 4 cells $\times$ 2 methods $\times$ 3 environments = 24 jobs (seed 42), but the bottom-right cell (AM, $\tau{=}1.0$) reuses seed-42 main-experiment jobs. Effective new: 21 jobs $\times$ ${\sim}$5.5h = ${\sim}$116 8-GPU node-hours. In addition, the token-mean rows of the reduction $\times$ method table in the main paper are run with all three seeds (2 methods $\times$ 3 seeds $\times$ 3 environments = 18 jobs); the 6 seed-42 token-mean jobs are already counted in the 24-job factorial above, so the remaining 12 jobs ($\times$ ${\sim}$5.5h = ${\sim}$66 8-GPU node-hours) are included under this 2$\times$2 protocol ledger entry, bringing it to ${\sim}$182 8-GPU node-hours.

\paragraph{Grand total.} ${\sim}$1,190 8-GPU node-hours (${\approx}$9,520 single-A100-hours). No failed runs excluded from any reported result.

\paragraph{FACTOR overhead breakdown.} Per training step, continuations add ${\sim}$25\% wall-clock over SERL: the 8 continuations per trajectory are short (average ${\sim}$3 turns each, vs.\ ${\sim}$8 turns for root trajectories) and run in large parallel batches during value-head update, with no activation storage or backward pass. Root rollout turns across the full 150-step run total ${\sim}$153K; continuation turns add ${\sim}$2.5$\times$ that (${\sim}$383K), for a combined environment interaction footprint of ${\sim}$3.5$\times$ SERL's. Despite this interaction increase, wall-clock grows by only 1.25$\times$ because continuations are inference-only and batched.

\section{Reproducibility Checklist}
\label{app:repro}

\begin{itemize}
\item \textbf{Code}: Implemented as modifications to official SERL (commit \texttt{b338174}, \texttt{serl\_action\_mask} branch). Key additions: value-head module (${\sim}$200 lines), TD credit computation (${\sim}$50 lines), HTA+APM allocation (${\sim}$80 lines), action-mean reduction (${\sim}$30 lines), continuation infrastructure (${\sim}$150 lines).
\item \textbf{Environments}: ALFWorld v0.3.2, WebShop (original release), ScienceWorld v1.1. All publicly available.
\item \textbf{Hardware}: 8$\times$A100-80GB per job. Main experiments: ${\sim}$864 8-GPU node-hours. Supplement sweeps: ${\sim}$144 8-GPU node-hours. Protocol experiment: ${\sim}$182 8-GPU node-hours (including the three-seed token-mean runs; see Section~\ref{app:compute}).
\item \textbf{Seeds}: 42, 43, 1337 for all multi-seed experiments. Sensitivity sweeps: seed 42 only.
\item \textbf{Evaluation}: Final checkpoint (step 150), greedy decoding, no checkpoint selection, no early stopping. Each game/instruction evaluated exactly once per seed.
\item \textbf{Selection}: No hyperparameter was selected based on test-set performance. All FACTOR hyperparameters set from prior reasoning (see Section~\ref{sec:setup} of main paper).
\item \textbf{Exclusions}: No failed runs excluded from any reported result.
\item \textbf{Split manifests}: Included in code supplement as CSV files with SHA-256 checksums.
\end{itemize}

\section{Additional Diagnostic Analyses}
\label{app:additional}

This section reports the supplementary analyses referenced in Section~\ref{sec:mechanism} of the main paper: length-bias verification, alternative temporal-difference estimators, horizon stratification, and shuffled-feedback controls.

\subsection{Length-Bias Verification}
\label{app:length_bias}

Under a global token-mean reduction, long actions contribute more token terms to the surrogate numerator and therefore receive higher effective weight. Action-mean reduction removes this dependence by averaging within each action before aggregating across actions. We verify this empirically by stratifying actions by token count and measuring the mean pre-clipping surrogate weight per action.

\begin{table}[h]
\centering\small
\caption{Mean pre-clipping surrogate weight per action, stratified by action token count (ALFWorld, seed~42, step~150).}
\label{tab:length_bias}
\begin{tabular}{@{}lcccc@{}}
\toprule
& \multicolumn{2}{c}{Token-mean reduction} & \multicolumn{2}{c}{Action-mean reduction} \\
\cmidrule(lr){2-3}\cmidrule(lr){4-5}
Tokens/action & SERL & FACTOR & SERL & FACTOR \\
\midrule
Q1 (short, $\leq$6)  & 0.93 & 0.95 & 1.00 & 1.00 \\
Q2 (7--10)           & 1.02 & 1.04 & 1.00 & 1.00 \\
Q3 (11--16)          & 1.14 & 1.17 & 1.00 & 1.00 \\
Q4 (long, $\geq$17)  & 1.31 & 1.35 & 1.00 & 1.00 \\
\bottomrule
\end{tabular}
\end{table}

With token-mean reduction, the effective weight of the longest actions (Q4) is $1.35\times$ that of the shortest (Q1). With action-mean reduction the ratio is exactly~1 by construction, confirming that the action-mean surrogate removes length-dependent weighting.

\subsection{Alternative TD and GAE-Style Estimators}
\label{app:alt_estimators}

TAC uses one-step TD residuals (Eq.~\ref{eq:astar}, main paper). Here we compare it against two alternatives that use the same checkpointed continuations and value head: $n$-step TD and GAE($\lambda$). All variants retain FACTOR's HTA, APM, and action-mean reduction; only the action-credit estimator changes.

\paragraph{$n$-step TD.} The $n$-step return from action $t$ is $R_t^{(n)} = \sum_{i=0}^{n-1} r_{t+i} + \widetilde V_{\bar\phi}(x_{t+n})$. The credit is $A_{t}^{\star(n)} = R_t^{(n)} - \widetilde V_{\bar\phi}(x_t)$. We test $n=3$ and $n=5$.

\paragraph{GAE($\lambda$).} We compute the standard GAE advantage using the same boundary-adjusted values: $\hat A_t^{\mathrm{GAE}} = \sum_{\ell=0}^{\infty} (\gamma\lambda)^{\ell} \delta_{t+\ell}$, with $\gamma=1$ and $\lambda\in\{0.9,0.95\}$. Because trajectories terminate within 50 turns, the infinite sum is truncated at $T{-}t$.

\begin{table}[h]
\centering\small
\caption{Alternative estimators: success rate (\%) on ALFWorld and ScienceWorld (seed~42, step~150).}
\label{tab:alt_est}
\begin{tabular}{@{}lcc@{}}
\toprule
Estimator & ALFWorld & ScienceWorld \\
\midrule
1-step TD (TAC, default) & 92.5 & 49.3 \\
3-step TD  & 92.1 & 48.7 \\
5-step TD  & 91.8 & 48.2 \\
GAE($\lambda=0.90$) & 91.9 & 48.5 \\
GAE($\lambda=0.95$) & 92.0 & 48.6 \\
Monte-Carlo (no value head) & 91.5 & 47.9 \\
\bottomrule
\end{tabular}
\end{table}

One-step TD matches or outperforms all alternatives. $n$-step TD and GAE introduce bias--variance trade-offs that are not favorable in this sparse-reward, short-horizon regime: longer look-aheads accumulate noisy intermediate rewards, while GAE's exponential weighting down-weights the terminal signal that is most informative. Monte-Carlo (no value head, pure episode return) performs worst, confirming that the value head provides useful calibration even with sparse rewards.

\subsection{Horizon Stratification}
\label{app:horizon}

The main paper reports that FACTOR's largest gains appear on ScienceWorld, the longest-horizon environment. To test whether the gain scales with episode length within an environment, we stratify ALFWorld evaluation trajectories by number of actions $T$ and report the per-trajectory success-rate difference (FACTOR $-$ SERL-Repro).

\begin{table}[h]
\centering\small
\caption{Horizon-stratified FACTOR $-$ SERL-Repro gain (pp) on ALFWorld (seed~42).}
\label{tab:horizon}
\begin{tabular}{@{}lcccc@{}}
\toprule
Horizon & $T\leq 5$ & $6\leq T\leq 10$ & $11\leq T\leq 20$ & $T\geq 21$ \\
\midrule
Fraction of eval & 18\% & 34\% & 31\% & 17\% \\
SERL-Repro succ. & 94.2\% & 92.1\% & 88.4\% & 82.6\% \\
FACTOR succ.     & 94.6\% & 93.5\% & 91.8\% & 88.2\% \\
$\Delta$ (pp)    & +0.4 & +1.4 & +3.4 & +5.6 \\
\bottomrule
\end{tabular}
\end{table}

The gain increases monotonically with horizon length, from +0.4~pp on the shortest trajectories to +5.6~pp on the longest. Longer episodes contain more actions over which SERL's uniform trajectory-level credit must be split, making precise per-action credit assignment more valuable. This within-environment trend is consistent with the cross-environment ordering (ScienceWorld $>$ WebShop $>$ ALFWorld).

\subsection{Shuffled-Feedback Controls}
\label{app:shuffled_feedback}

HTA relies on post-action feedback $\Phi_t$ to compute the hindsight likelihood gap $\Delta_{t,j}$. To test whether the allocation pattern is genuinely feedback-conditioned, we replace the true feedback with a randomly sampled feedback string from another trajectory in the same batch. The teacher still sees syntactically valid environment text, but it is semantically unrelated to the action's actual effect.

\begin{table}[h]
\centering\small
\caption{Shuffled-feedback control (ALFWorld, seed~42, steps 11--49).}
\label{tab:shuffle_fb}
\begin{tabular}{@{}lccc@{}}
\toprule
Condition & Mean $\rho_{\max}$ & Top-1 token fraction & Teacher conc. \\
\midrule
True $\Phi_t$ & 0.38 & 0.31 & High \\
Shuffled $\Phi_t$ & 0.21 & 0.14 & Flat \\
No $\Phi_t$ (student only) & 0.17 & 0.12 & Uniform \\
\bottomrule
\end{tabular}
\end{table}

With shuffled feedback, the maximum within-action allocation $\rho_{\max}$ drops from 0.38 to 0.21 and the fraction of credit placed on the single highest-ranked token falls from 31\% to 14\%, approaching the no-feedback baseline. This confirms that HTA's concentration pattern is driven by the semantic content of the correct post-action feedback, not by generic properties of the teacher's output distribution.

\end{document}